\documentclass[sigconf]{acmart}

\AtBeginDocument{%
  \providecommand\BibTeX{{%
    \normalfont B\kern-0.5em{\scshape i\kern-0.25em b}\kern-0.8em\TeX}}}

\setcopyright{acmcopyright}
\copyrightyear{2018}
\acmYear{2018}
\acmDOI{10.1145/1122445.1122456}

\acmConference[Woodstock '18]{Woodstock '18: ACM Symposium on Neural
  Gaze Detection}{June 03--05, 2018}{Woodstock, NY}
\acmBooktitle{Woodstock '18: ACM Symposium on Neural Gaze Detection,
  June 03--05, 2018, Woodstock, NY}
\acmPrice{15.00}
\acmISBN{978-1-4503-XXXX-X/18/06}

\acmSubmissionID{157}

\usepackage{hyperref}
\usepackage{url}
\usepackage{enumitem}
\usepackage{wrapfig}
\usepackage{booktabs}       
\usepackage{graphicx}
\usepackage{subcaption}     
\usepackage{algorithm}
\usepackage{algpseudocode}
\usepackage{multirow}
\usepackage{xcolor}
\usepackage{amsthm}
\usepackage{amsmath}
\usepackage{marginnote}
\usepackage{natbib}
\usepackage{xfrac}

\definecolor{myblue2}{HTML}{4682B4}
\definecolor{myred}{HTML}{FF5733}

\usepackage{amsmath,amsfonts,bm}

\def\eqref#1{equation~\ref{#1}}

\def\1{\bm{1}}

\DeclareMathAlphabet{\mathsfit}{\encodingdefault}{\sfdefault}{m}{sl}
\SetMathAlphabet{\mathsfit}{bold}{\encodingdefault}{\sfdefault}{bx}{n}

\begin{document}

\title{REX: Revisiting Budgeted Training with an Improved Schedule}

\author{John Chen}
\email{johnchen@rice.edu}
\affiliation{%
  \institution{Rice University}
  \city{Houston}
  \state{Texas}
  \country{USA}
}
\author{Cameron Wolfe}
\email{wolfe.cameron@rice.edu}
\affiliation{%
  \institution{Rice University}
  \streetaddress{1601 Rice Blvd}
  \city{Houston}
  \state{Texas}
  \country{USA}
}
\author{Anastasios Kyrillidis}
\email{anastasios@rice.edu}
\affiliation{%
  \institution{Rice University}
  \streetaddress{1601 Rice Blvd}
  \city{Houston}
  \state{Texas}
  \country{USA}
}

\renewcommand{\shortauthors}{Chen et al.}

\begin{abstract}
   Deep learning practitioners often operate on a computational and monetary budget.
    Thus, it is critical to design optimization algorithms that perform well under any budget.
    The linear learning rate schedule is considered the best budget-aware schedule \cite{li2020budgeted}, as it outperforms most other schedules in the low budget regime.
    On the other hand, learning rate schedules --such as the \texttt{30-60-90} step schedule-- are known to achieve high performance when the model can be trained for many epochs.
    Yet, it is often not known a priori whether one's budget will be large or small; thus, the optimal choice of learning rate schedule is made on a case-by-case basis. 
    In this paper, we frame the learning rate schedule selection problem as a combination of $i)$ selecting a profile (i.e., the continuous function that models the learning rate schedule), and $ii)$ choosing a sampling rate (i.e., how frequently the learning rate is updated/sampled from this profile).
    We propose a novel profile and sampling rate combination called the Reflected Exponential (REX) schedule, which we evaluate across seven different experimental settings with both SGD and Adam optimizers.
    REX outperforms the linear schedule in the low budget regime, while matching or exceeding the performance of several state-of-the-art learning rate schedules (linear, step, exponential, cosine, step decay on plateau, and OneCycle) in both high and low budget regimes.
    Furthermore, REX requires no added computation, storage, or hyperparameters.
\end{abstract}



\begin{teaserfigure}
  \centering
    \begin{subfigure}[b]{0.40\linewidth}
    \includegraphics[width=\linewidth]{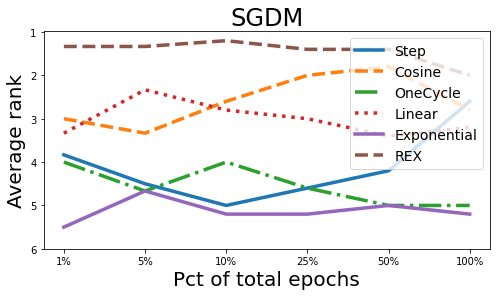}
    \label{fig:SummarySGDM}
  \end{subfigure}
  \begin{subfigure}[b]{0.40\linewidth}
    \includegraphics[width=\linewidth]{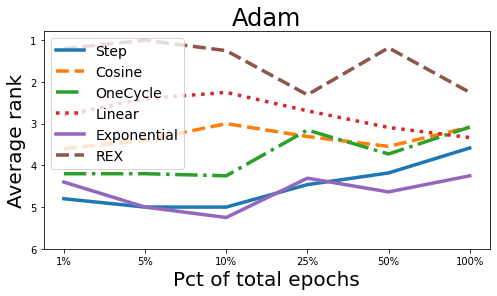}
    \label{fig:SummaryAdam}
  \end{subfigure}\vspace{-0.6cm}
  \caption{We summarize all 82 experimental settings, including image classification, object detection, and natural language processing. We plot the average ranked performance of the considered learning rate schedules, where 1 is the best and 6 is the worst, against the training budget, for the momentum SGD (SGDM) and Adam optimizers. The maximum epochs (100\%) is determined from the literature and verified to achieve previously reported results. Each \% of total epochs is an independent run. The schedules are adjusted for each setting to maintain the same profile (e.g. the linear schedule decays the learning rate linearly to 0 regardless of the \% of total epochs). For smaller epochs, the linear schedule performs well while the performance of the step schedule in higher epochs does not carry over. \textit{The proposed REX schedule outperforms all methods in comparison, in both high and low epochs.}}
  \Description{}
  \label{fig:teaser}
\end{teaserfigure}

\maketitle

\begin{table*}
\centering
\caption{\small Performance of different schedules, ranked according to the \% of Top-1 or Top-3 finishes, out of a total of 28 experiments. Top-1 (Top-3) refers to the best (best-3) performance for a particular model/dataset/base optimizer/epoch setting. Low (high) budget includes 1\%, 5\%, and 10\% (25\%, 50\%, and 100\%) of the full epochs. The Decay on Plateau variant is aggregated into the Step Schedule method where we take the max performance for each setting. 
}
\begin{tabular}{ccccccc} 
\toprule
     &  \multicolumn{2}{c}{Low budget ($<$25\%)} & \multicolumn{2}{c}{High budget ($\geq$25\%)} & \multicolumn{2}{c}{Overall} \\ \midrule
    Method & Top-1 & Top-3 & Top-1 & Top-3 & Top-1 & Top-3 \\ \midrule 
    None & 0\% & 0\% & 2\% & 10\% & 1\% & 5\% \\
    Exp decay \cite{tensorflow2015-whitepaper, paszke2017automatic} & 5\% & 7\% & 5\% & 14\% & 5\% & 11\% \\
    OneCycle \cite{smith20181cycle} & 15\% & 49\% & 12\% & 40\% & 13\% & 45\% \\
    Linear Schedule \cite{tensorflow2015-whitepaper, paszke2017automatic} & 10\% & 78\% & 12\% & 62\% & 11\% & 70\% \\
    Step Schedule \cite{he2016deep} & 2\% & 12\% & 7\% & 38\% & 5\% &  25\% \\
    Cosine Schedule \cite{loshchilov2017sgdr} & 2\% & 66\% & 10\% & 62\% & 6\% & 64\% \\ \midrule
    \textsc{REX} & \textcolor{myred}{\textbf{73\%}} & \textcolor{myred}{\textbf{95\%}} & \textcolor{myred}{\textbf{67\%}} & \textcolor{myred}{\textbf{88\%}} & \textcolor{myred}{\textbf{70\%}} & \textcolor{myred}{\textbf{92\%}}\\
    \bottomrule
\end{tabular}
\label{tablepctresults}
\end{table*}

\section{Introduction}
While hardware has consistently improved \cite{Sze_2017, Shawahna_2019}, the cost of training deep neural networks (DNNs) has continued to increase due to growth in the size of models and datasets \cite{krizhevsky2012imagenet, devlin2019bert, brown2020language, chen2020big}. 
One key component of the cost is the need to tune the hyperparameters of the model \cite{Yang_2020}. 
Outside of the largest companies in the field, most practitioners have to trade-off the number of epochs with the number of experimental trials. 
Whilst the community has generally agreed that, for example, 90 epochs is a reasonable training length for a ResNet-50 architecture on ImageNet \cite{he2016deep, huang2017densely, zagoruyko2016wide}, there simply may not be sufficient monetary budget to perform such extensive training for certain projects.
Further, it is generally not easy to predict the number of epochs required to maximize the performance of the model apriori, particularly if the input data may be continually changing. 
\textit{Thus, it is important to consider the optimization of DNNs for a diverse range of budgets.}

Stochastic Gradient Descent (SGD) with momentum and Adam are two of the most widely used optimizers for DNNs \cite{he2016deep, huang2017densely, zagoruyko2016wide, devlin2019bert, brown2020language, redmon2018yolov3}. 
Whether the task is image classification, object detection, or fine-tuning in natural language processing, both optimizers must be combined with some form of learning rate decay to achieve good performance \cite{he2016deep, huang2017densely, zagoruyko2016wide, devlin2019bert, brown2020language, redmon2018yolov3} (see Tables \ref{tablern20cifar10}-\ref{tablebertgluecomp}). 
The aforementioned tasks are arguably the most widely used applications of deep learning.\footnote{There are some cases in which learning rate decay is not always useful, such as for Generative Adversarial Networks \cite{goodfellow2014generative, arjovsky2017wasserstein}, but this is generally a small proportion of all deep learning activities.}

The learning rate schedule is particularly important in the budgeted training setting.
Moreover, of the widely used schedules, the best learning rate schedule for a small number of epochs is generally not the best for a large number of epochs (see Tables \ref{tablern20cifar10}-\ref{tablebertgluecomp}). 
This is a significant challenge, since it is difficult to know apriori if the current budget lies in the high or low budget regime. 
This raises two questions: \textit{Can we close the budget-induced gap in the performance of existing learning rate schedules? And, if this is not possible, is there a learning rate schedule that performs well in both low and high budget regimes?}

We answer both questions through a novel lens. 
We decompose the problem of selecting a learning rate schedule as a two-part process of $i)$ selecting a profile and $ii)$ selecting a sampling rate.
The \textit{profile} is the function that models the learning rate schedule, and the \textit{sampling rate} is how frequently the learning rate is updated, based on this profile. 
In this view, we $i)$ analyze existing schedules, $ii)$ propose a novel profile and sampling rate combination, and $iii)$ benchmark the performance of numerous schedules. 
We also demonstrate it is possible to boost the performance of existing learning rate schedules by introducing a hyperparameter that delays the commencement of the decay schedule. 
However, because adding an extra hyperparameter is prohibitive in the budgeted setting, we also propose a new schedule, REX, which performs at a state-of-the-art level for both low and high budgets across a large variety of settings without the extra hyperparameter tuning. 

Specifically, our contributions are as follows: 
\begin{itemize}[leftmargin=*]
    \item We pose learning rate schedules as the combination of a profile and a sampling rate and identify that there is no optimal profile for all sampling rates. 
    Namely, we show that no existing, popular learning rate schedule achieves state-of-the-art performance in both high and low budget regimes.
    \item We propose a new profile and sampling rate combination. We find that carefully tuning the start of the learning rate decay for existing schedules can result in significant performance improvements in both high and low budget regimes. However, this introduces an extra hyperparameter, which is prohibitive for budget-limited practitioners.  
    Our proposed schedule can be understood as an interpolation between the linear schedule and the delayed variants.
    \item Our proposed schedule, REX, is based on observations of the above, and we validate its state-of-the-art performance across seven settings, including image classification, object detection, and natural language processing. 
\end{itemize}
\textit{Our goal is to introduce an easy-to-use, state-of-the-art learning rate schedule with no extra hyperparameters that performs well in all budget regimes and can be easily implemented and adopted.}

\begin{figure*}
  \centering
    \begin{subfigure}[b]{0.24\linewidth}
    \includegraphics[width=\linewidth]{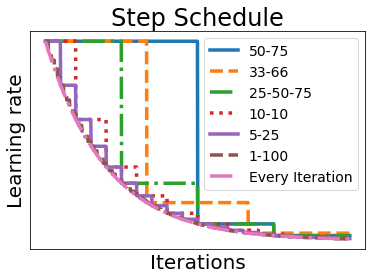}
    \label{fig:stepSampled}
  \end{subfigure}
  \begin{subfigure}[b]{0.24\linewidth}
    \includegraphics[width=\linewidth]{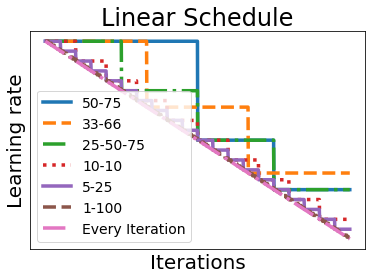}
    \label{fig:linearSampled}
  \end{subfigure} 
  \begin{subfigure}[b]{0.24\linewidth}
    \includegraphics[width=\linewidth]{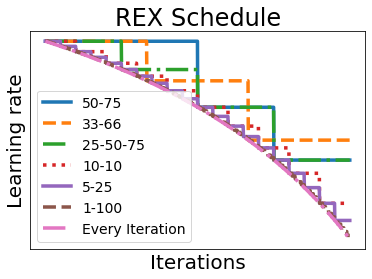}
    \label{fig:REXSampled}
  \end{subfigure} 
  \begin{subfigure}[b]{0.24\linewidth}
    \includegraphics[width=\linewidth]{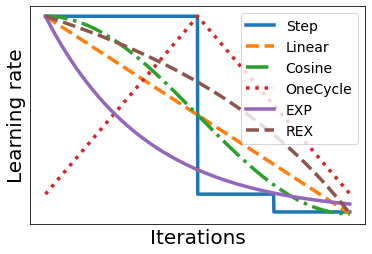}
    \label{fig:allSchedules}
  \end{subfigure}
  \caption{\small Popular schedules with various sampling rates. \texttt{50-75} refers to sampling once at 50\% and 75\% of total epochs. Similarly for \texttt{33-66} and \texttt{25-50-75}. \texttt{10-10} refers to sampling once every 10\% of total epochs. Similarly for \texttt{5-25} and \texttt{1-100}. Every iteration is the maximum sampling rate. Left: Step schedule. Left Middle: Linear Schedule. Right Middle: REX Schedule. Right: Schedules with their usual sampling rate. 
  }
  \label{fig:schedulesAndRate}
\end{figure*}

\section{Related Works}
There have been many works related to tuning the learning rate. 
There is a connection between learning rate and momentum \citep{yuan2016sgdequivalence}, and there are methods which alter the momentum \citep{sutskever2013importance, zhang2017yellowfin, o2015adaptive, lucas2018aggregated, chen2020demon}. 
There is also a connection between learning rate and batch sizes \citep{smith2017increasebatch, you2017large, goyal2018accurate}. 
The most popular learning rate tuning mechanisms fall into two categories: Automatically tuning the learning rate on a per-weight basis and decaying the learning rate globally.

Many adaptive learning rate optimizers have been proposed. Modern learning rate adaptive methods began with AdaGrad \citep{duchi2011adaptive}, which was shown to have good convergence properties, especially in the sparse gradient setting. 
AdaDelta \citep{zeiler2012adadelta} was proposed to fix a units issue with AdaGrad. 
RMSprop \citep{hinton2012neural} employed a running estimate of the second moment to resolve the strictly decreasing learning rate of AdaGrad. 
The most popular adaptive learning rate optimizer is Adam \citep{kingma2014adam} and its variants \citep{loshchilov2017fixing, liu2020variance}. 
\textit{Yet, in practice, adaptive learning rate algorithms perform the best when coupled with a learning rate schedule} \cite{devlin2019bert, liu2020variance}.

In deep learning, the step schedule was widely used in early computer vision work \citep{krizhevsky2012imagenet, he2016deep, huang2017densely}. 
This was often combined with SGD with Momentum to achieve state-of-the-art results \citep{he2016deep, zagoruyko2016wide, huang2017densely, redmon2016look}. 
In Natural Language Processing, AdamW \citep{loshchilov2017fixing} is often paired with a cosine or linear learning rate decay for training and fine-tuning transformers \citep{wolf2020huggingfaces}. 
The aforementioned schedules are widely available and implemented in the most popular software \citep{wolf2020huggingfaces, tensorflow2015-whitepaper, paszke2017automatic}, in addition to the exponential decay schedule, OneCycle \citep{smith20181cycle}, cosine decay with restarts \citep{loshchilov2017sgdr} and others \citep{smith2017cyclical}. 
While some schedules may be preferred for achieving state-of-the-art results, it has been suggested that the linear schedule is most suitable for the low budget scenario \citep{li2020budgeted}, which may be of more relevance to practitioners.

\section{Budgeted Training: Profiles and Sampling Rates}
\textbf{Challenges in adapting learning rate schedules to the budgeted setting.} 
The primary hyperparameter in DNN optimization is the initial learning rate. 
While good heuristics often exist for tuning common hyperparameters, such as setting momentum $\beta = 0.9$ or setting a \texttt{30-60-90} learning rate schedule \cite{zagoruyko2016wide, huang2017densely, hu2017squeeze}, 
the initial learning rate remains to be tuned. 
However, in the budgeted training setting, the learning rate schedule turns into a hyperparameter.
Adapting, for example, the \texttt{30-60-90} rule for Image Classification or Object Detection is not straightforward, and naively following the same rules for a smaller number of epochs results in sub-optimal results (see Step Schedule in low epoch settings in Tables \ref{tablern20cifar10}-\ref{tablebertgluecomp}). 
Additionally, following the \texttt{50-75} rule \cite{he2016deep} on \texttt{RN20-CIFAR10} 
for a training budget that is 1\% of the usual total epochs can result a 5\% absolute error gap with the best-performing schedule. 
We assume that, in the budgeted training setting, the number of epochs is still pre-defined, but can be significantly less than the usual total epochs.

\textbf{Profiles and sampling rates.} 
To formalize the process of identifying a good learning rate schedule, we decompose the learning rate schedule as a combination of a profile curve and a sampling rate on that curve.
The \textit{profile} is the function that models the learning rate schedule and dictates the general curve of the learning rate schedule.
In most --but not all \cite{Li2020An}-- applications, this function starts at a high initial value and ends near zero. 
The \textit{sampling rate} is how frequently the learning rate is updated and dictates the smoothness of the curve. 
At one extreme, the linear learning rate schedule, and many others, samples from the profile at each iteration, and at the other extreme the step learning rate schedules samples only twice or thrice across the entire training procedure. 
For example, the \texttt{50-75} step schedule can be approximated as sampling twice from a particular, exponentially-decaying profile. 
See Figure \ref{fig:schedulesAndRate} for some examples of schedules with their associated profile and sampling rates. 

\textbf{Lack of an optimal profile.} 
While there may be limited motivation to pick a particular sampling rate, this introduces an interesting question: \textit{Does there exist an optimal profile for all reasonable sampling rates?} 
In Table \ref{tableNoBestProfileresults}, we benchmark three profiles: $i)$ the \texttt{50-75} step schedule \cite{he2016deep} approximated as a tuned exponentially decaying profile 
; $ii)$ the linear profile \cite{tensorflow2015-whitepaper,paszke2017automatic}; and, $iii)$ the REX profile proposed in this paper (to be defined in the next subsection).
These three profiles represent smoothly-decaying learning rate schedules with varying curvatures. 
We find that different profiles perform best for different sampling rates.
\textit{The approximated Step schedule profile performs best with low sampling rates, while the linear and REX profiles perform best with high sampling rates.}
Furthermore, \textit{the approximated Step schedule profile performs worst for a small and medium number of epochs and best for a high number of epochs.} 
\textbf{The REX profile performs best for a small and medium number of epochs.} 
While the Step schedule is consistently used to achieve state-of-the-art results in Computer Vision \cite{he2016deep, zagoruyko2016wide, huang2017densely, hu2017squeeze, redmon2016look, he2018mask}, it does not translate directly to lower epoch settings.

\begin{table*}
\centering
\caption{We demonstrate learning rate schedules and sampling rates on \texttt{RN20-CIFAR10-SGDM} (Top) and \texttt{RN38-CIFAR10-SGDM} (Bottom) \cite{he2016deep}, holding the learning rate constant. There is no best profile for all sampling rates. Each profile excels at one end of the spectrum. \texttt{50-75} \cite{he2016deep} refers to sampling once at 50\% and 75\% of total epochs. Similarly for \texttt{33-66} and \texttt{25-50-75}. \texttt{10-10} refers to sampling once every 10\% of total epochs. Similarly for \texttt{5-25} and \texttt{1-100}. Every iteration is the maximum sampling rate. 
}
\begin{tabular}{c|ccc|ccc|ccc} \toprule
     \texttt{RN20-CIFAR10-SGDM} & \multicolumn{3}{c|}{15 Epochs} & \multicolumn{3}{c|}{75 Epochs}  & \multicolumn{3}{c|}{300 Epochs} \\ \midrule
    Sampling Rate & Step  & Linear & REX &  Step & Linear & REX &  Step & Linear & REX \\ \midrule
    \texttt{50-75} & 14.48 & 16.96 & 20.79 & 9.44 & 12.42 & 18.05 & \textcolor{red}{\textbf{7.32}} & 10.15 & 12.41 \\
    \texttt{33-66} & 17.89 & 25.80 & 24.45 & 9.72 & 13.38 & 15.98 & 7.93 & 11.90 & 11.43 \\
    \texttt{25-50-75} & 16.52 & 18.77 & 26.13 & 9.73 & 12.31 & 12.59 & 8.46 & 8.26 & 12.31  \\
    \texttt{10-10} & 17.98 & 16.35 & 16.48 & 10.41 & 9.40 & 11.17 & 8.67 & 8.26 & 8.24 \\
    \texttt{5-25} & 18.87 & 13.83 & 15.17 & 9.79 & 8.94 & 9.22 & 8.85 & 8.24 & 8.50 \\
    \texttt{1-100} & 18.53 & 13.91 & \textbf{13.34} & 10.61 & \textbf{8.72} & \textbf{8.60} & 9.20 & 7.97 & 7.74 \\
    Every Iteration & 19.19 & \textbf{13.09} & \textcolor{red}{\textbf{12.86}} & 9.97 & 8.89 & \textcolor{red}{\textbf{8.37}} & 9.24 & \textbf{7.62} & \textbf{7.52} \\ 
    \bottomrule \toprule
    \texttt{RN38-CIFAR10-SGDM}& \multicolumn{3}{c|}{15 Epochs} & \multicolumn{3}{c|}{75 Epochs}  & \multicolumn{3}{c|}{300 Epochs} \\ \midrule
    Sampling Rate & Step  & Linear & REX &  Step & Linear & REX &  Step & Linear & REX \\ \midrule
    \texttt{50-75} & 13.57 & 17.31 & 18.47 & 7.59 & 12.89 & 14.38 & 6.66 & 10.07 & 9.37 \\
    \texttt{33-66} & 14.96 & 19.16 & 18.71 & 7.74 & 13.64 & 17.57 & 6.70 & 11.53 & 11.30 \\
    \texttt{25-50-75} & 15.69 & 14.18 & 19.77 & 7.99 & 9.10 & 15.07 & 6.73 & 7.59 & 8.44 \\
    \texttt{10-10} & 16.58 & 13.34 & 14.46 & 7.87 & 8.33 & 9.75 & 7.60 & 6.48 & 6.50 \\
    \texttt{5-25} & 17.16 & 12.63 & \textbf{11.71} & 8.40 & 7.42 & 7.13 & 8.79 & 6.18 & 6.41\\
    \texttt{1-100} & 17.20 & 11.93 & \textbf{11.13} & 8.54 & \textbf{7.06} & 7.17 & 9.11 & \textbf{6.12} & 6.17 \\
    Every Iteration & 17.97 & 12.11 & \textcolor{red}{\textbf{10.95}} & 8.72 & \textbf{7.10} & \textcolor{red}{\textbf{6.86}} & 9.31 & \textcolor{red}{\textbf{5.89}} & \textbf{6.09} \\ \bottomrule
\end{tabular}
\label{tableNoBestProfileresults}
\end{table*}

\textbf{A new profile.} 
Since there is no profile which performs optimally across sampling rates, it remains to ask if there is a profile and sampling rate combination that results in strong performance in both low and high epoch settings. 
Therefore, we propose the Reflected Exponential (REX) profile; see Figure \ref{fig:schedulesAndRate}. 
REX is an alternative to the linear and exponential profile, and we find that REX has stronger empirical performance in the budgeted setting.
REX performs best with a per-iteration sampling rate, similar to the linear schedule. 
We evaluate the performance of REX extensively in following sections.

\begin{figure*}
  \centering
    \begin{subfigure}[b]{0.24\linewidth}
    \includegraphics[width=\linewidth]{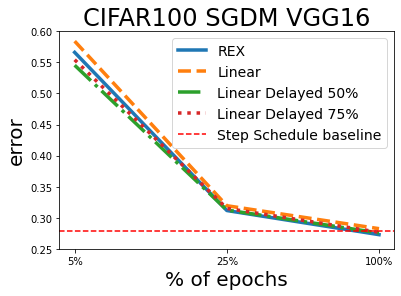}
    \label{fig:emp1}
  \end{subfigure}
  \begin{subfigure}[b]{0.24\linewidth}
    \includegraphics[width=\linewidth]{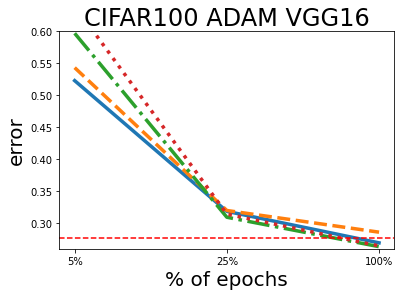}
    \label{fig:emp2}
  \end{subfigure} 
  \begin{subfigure}[b]{0.24\linewidth}
    \includegraphics[width=\linewidth]{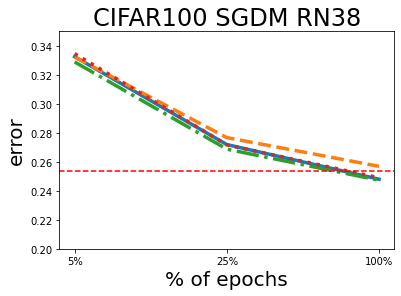}
    \label{fig:emp3}
  \end{subfigure} 
  \begin{subfigure}[b]{0.24\linewidth}
    \includegraphics[width=\linewidth]{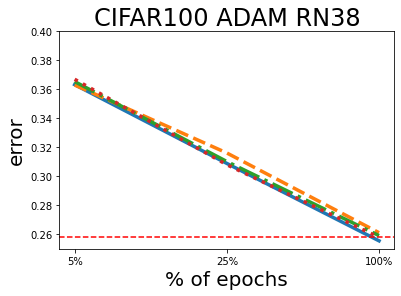}
    \label{fig:emp4}
  \end{subfigure}
  \caption{\small REX, linear, and delayed linear schedules. Left: \texttt{VGG16-CIFAR100-SGDM}. Left Middle: \texttt{VGG16-CIFAR100-ADAM}. Right Middle: \texttt{RN38-CIFAR100-SGDM}. Right: \texttt{RN38-CIFAR100-ADAM}. The red dashed line represents the error of the step schedule for that setting trained with 100\% of the epochs. Linear Delayed X\% refers to delaying the linear decay till X\% of the total epochs have passed, before decaying linearly to 0. For example, in the left-middle plot, for small \% of epochs, REX outperforms the linear schedule, which outperforms the delayed variants. However, for large epochs, the linear schedule is unable to achieve the state-of-the-art performance of the step schedule, while REX and the delayed linear schedules are able to surpass the step schedule. 
  }
  \label{fig:empiricalMotivation}
\end{figure*}

We also motivate REX with the empirical observation that the linear schedule can be improved in some cases by delaying the onset of the decay, i.e., holding the initial learning rate constant until \texttt{XX}\% of the budget, and then linearly decaying the learning rate to 0; see Figure \ref{fig:empiricalMotivation}. 
In particular, it appears that performance can be improved with such delay in the high epoch regime, but this strategy is less effective with fewer epochs.
However, \textit{the exact onset of the delay introduces an additional hyperparameter}. 
REX can be understood as an interpolation between a linear schedule and a delayed linear schedule without additional hyperparameters.
Furthermore, REX generally outperforms the linear schedule, which has been previously suggested as the best budgeted schedule \cite{li2020budgeted}, for small and large epochs.

It appears that certain schedules have reasonable performance across sampling rates, while others have poor or state-of-the-art performance depending on the sampling rate.
If the sampling rate is unknown or there is a particular reason to select a low sampling rate, the approximated step profile appears to be the best choice. 
However, in most applications, the sampling rate is a choice by the practitioner.
Since the REX profile with a per-iteration sampling rate generally performs the best, there may be limited motivation to use alternative schedules.

\begin{table*}[!htp]
\centering
\caption{\small Summary of experimental settings.}\label{experimentSummary}
\begin{tabular}{l|l|l|l} 
\toprule
    Experiment short name & Model & Dataset & Maximum Epochs \\ \midrule
    \texttt{RN20-CIFAR10}  & ResNet20  & CIFAR10 & 300 \cite{he2016deep} 
    \\
    \texttt{RN50-IMAGENET} & ResNet50 & ImageNet & 90 \cite{huang2017densely} \\
    \texttt{VGG16-CIFAR100} & VGG-16 & CIFAR100 & 300 \cite{he2016deep} \\
    \texttt{WRN-STL10} & Wide ResNet 16-8 & STL10 & 200 \cite{chang2017reversible} \\
    \texttt{VAE-MNIST} & VAE & MNIST & 200 \cite{yeung2017tackling} \\
    \texttt{YOLO-VOC} & YOLOv3 & Pascal VOC & 50 \cite{tripathi2016context} \\
    \texttt{BERT$_{BASE}$-GLUE} & BERT (Pre-trained) & GLUE (9 tasks) & 3 \cite{devlin2019bert} \\
 \bottomrule
\end{tabular}
\end{table*}
\begin{table*}[t]
\centering
\caption{ \texttt{RN20-CIFAR10}. The number of epochs was predefined before the execution of the algorithms. \textcolor{myred}{\textbf{Bold red}} indicates Top-1 performance, \textbf{black bold} is Top-3.}
\begin{tabular}{c|cccccc}
\toprule
    SGDM & 1\% & 5\% & 10\% & 25\% & 50\% & 100\% \\ \midrule
    + Step Schedule & 32.14 $\pm$ .34 & 14.94 $\pm$ .27 & 11.80 $\pm$ .11& \textbf{8.82} $\pm$ .25 & 8.43 $\pm$ .07 & \textcolor{myred}{\textbf{7.32}} $\pm$ .14  \\
    + Cosine Schedule & \textbf{28.49} $\pm$ .25 & \textbf{13.05} $\pm$ .17 & \textbf{10.62} $\pm$ .29 & \textbf{8.80} $\pm$ .08 & \textbf{8.10} $\pm$ .13 & 7.78 $\pm$ .14\\
    + OneCycle & 40.14 $\pm$ 2.62 & 18.93 $\pm$ 1.85 & 12.74 $\pm$ .36 & 10.83 $\pm$ .25 & 9.23 $\pm$ .19 & 8.42 $\pm$ .12\\  
    + Linear Schedule & \textbf{28.70} $\pm$ 1.13 & \textbf{13.09} $\pm$ .13 & \textbf{10.85} $\pm$ .15 &  9.03 $\pm$ .24 & \textbf{8.15} $\pm$ .12 & \textbf{7.62} $\pm$ .12\\ 
    + Decay on Plateau & 41.98 $\pm$ 3.20 & 25.93 $\pm$ .45 & 11.29 $\pm$ .35 & 9.05 $\pm$ .07 & 8.26 $\pm$ .07 & 7.97 $\pm$ .14\\ 
    + Exp decay & 31.31 $\pm$ 1.34 & 14.85 $\pm$ .38 & 11.56 $\pm$ .22 & 9.55 $\pm$ .09 & 9.20 $\pm$ .13 &  7.82 $\pm$ .05\\ \midrule
    + REX & \textcolor{myred}{\textbf{27.94}} $\pm$ .46 & \textcolor{myred}{\textbf{12.86}} $\pm$ .27 & \textcolor{myred}{\textbf{10.23}} $\pm$ .13 & \textcolor{myred}{\textbf{8.37}} $\pm$ .09 & \textcolor{myred}{\textbf{7.52}} $\pm$ .24 & \textbf{7.52} $\pm$ .05 \\ \midrule
    Adam & 42.10 $\pm$ 2.71 & 23.01 $\pm$ 1.10 & 16.58 $\pm$ .18 & 13.63 $\pm$ .22 & 11.90 $\pm$ .06 & 11.94 $\pm$ .06 \\
    + Step Schedule & 30.72 $\pm$ .16 & 15.41 $\pm$ .26 & 12.20 $\pm$ .11 & 10.47 $\pm$ .10 & \textbf{8.75} $\pm$ .17 & \textbf{8.55} $\pm$ .05  \\  
    + Cosine Schedule & \textbf{29.20} $\pm$ .24 & \textbf{14.31} $\pm$ .28 & \textbf{11.45} $\pm$ .27 & \textbf{9.56} $\pm$ .12 & 9.15 $\pm$ .12 & 8.93 $\pm$ .07 \\    
    + OneCycle & 37.17 $\pm$ 2.49 & 16.16 $\pm$ .19 & 14.11 $\pm$ .57 & 10.33 $\pm$ .20 & 9.87 $\pm$ .12 & 9.03 $\pm$ .18\\  
    + Linear Schedule & \textbf{28.99} $\pm$ .37 & \textbf{14.08} $\pm$ .34 & \textbf{10.97} $\pm$ .19 & \textcolor{myred}{\textbf{9.25}} $\pm$ .12 & 9.20 $\pm$ .22 & 8.89 $\pm$ .05 \\ 
    + Decay on Plateau & 43.40 $\pm$ 4.57 & 22.21 $\pm$ .96 & 13.46 $\pm$ .38 & 9.71 $\pm$ .39 & \textbf{8.92} $\pm$ .18 & 8.80 $\pm$ .11\\ 
    + Exp decay & 31.87 $\pm$ .59 & 15.82 $\pm$ .06 & 12.91 $\pm$ .21 & 10.48 $\pm$ .15 & 9.24 $\pm$ .16 &  \textbf{8.53} $\pm$ .07  \\ \midrule
    + REX & \textcolor{myred}{\textbf{27.64}} $\pm$ .02 & \textcolor{myred}{\textbf{13.96}} $\pm$ .16 & \textcolor{myred}{\textbf{10.88}} $\pm$ .05 & \textbf{9.44} $\pm$ .22 & \textcolor{myred}{\textbf{8.72}} $\pm$ .24 & \textcolor{myred}{\textbf{8.18}} $\pm$ .15 \\
 \bottomrule
\end{tabular}
\label{tablern20cifar10}
\end{table*}

\section{Results}
In this section we present results in all seven experimental settings given in Table \ref{experimentSummary}, including image classification, image generation, object detection and natural language processing. 
For fair evaluation in the budgeted training scenario, only the learning rate is tuned in multiples of 3 for each schedule, setting, and number of epochs. 
All reported metrics are averaged across three separate trials.
We run all settings at 1\%, 5\%, 10\%, 25\%, 50\%, and 100\% of maximum epochs, representing both low and high budgets. 
In each setting, the learning rate schedule is concerned only with the total epochs for that run, e.g., the linear schedule will decay linearly to 0 regardless if the budget is 1\% or 100\% of the maximum epochs. 
For \texttt{BERT$_{BASE}$-GLUE}, results are given for 1 run and at $\sfrac{1}{3}$, $\sfrac{2}{3}$, and $\sfrac{3}{3}$ of total epochs. 
The maximum total epochs is determined from commonly used epochs in the literature, and validated to achieve the reported score in the literature. 
The maximum epochs is given in Table \ref{experimentSummary}. The goal is to demonstrate performance in both the low and high budget regime across a range of common applications to instill confidence that the proposed schedule will work ``in the wild''. We use a model-dataset-optimizer notation, e.g. RN20-CIFAR10-SGDM means a ResNet20 model trained on CIFAR10 with momentum SGD.

\begin{table*}[t]
\centering
\caption{ \texttt{WRN-STL10}. The number of epochs was predefined before the execution of the algorithms. \textcolor{myred}{\textbf{Bold red}} indicates Top-1 performance, \textbf{black bold} is Top-3.}
\begin{tabular}{c|cccccc}
\toprule
    SGDM & 1\% & 5\% & 10\% & 25\% & 50\% & 100\% \\ \midrule
    + Step Schedule & 60.09 $\pm$ 1.15 & 38.12 $\pm$ .32 & 33.86 $\pm$ .10 & 22.42 $\pm$ .56 & \textbf{17.20} $\pm$ .35 & \textcolor{myred}{\textbf{14.51}} $\pm$ .26 \\
    + Cosine Schedule   & \textbf{57.81} $\pm$ 1.05 & 37.42 $\pm$ .29 & \textbf{27.51} $\pm$ .25 &  \textbf{20.03} $\pm$ .26 & \textbf{17.02} $\pm$ .24 & 14.66 $\pm$ .25 \\    
    + OneCycle   & 58.75 $\pm$ .76 & \textbf{36.90} $\pm$ .37 & {\textbf{26.97}} $\pm$ .27 &  21.67 $\pm$ .27 & 19.69 $\pm$ .21 & 19.00 $\pm$ .42 \\  
    + Linear Schedule  & \textbf{58.74} $\pm$ 1.26 & {\textbf{34.81}} $\pm$ .40 & 28.17 $\pm$ .64 & \textcolor{myred}{\textbf{19.54}} $\pm$ .20 & 17.39 $\pm$ .24 & \textbf{14.58} $\pm$ .18 \\ 
    + Decay on Plateau  & 59.64 $\pm$ .92 & 37.64 $\pm$ 1.44 & 36.94 $\pm$ 1.96 &  21.05 $\pm$ .27 & 17.83 $\pm$ .39 & 15.16 $\pm$ .36 \\ 
    + Exp decay   & 60.21 $\pm$ .77 & 38.94 $\pm$ 1.08 & 34.11 $\pm$ .77 &  22.65 $\pm$ .49 & 20.60 $\pm$ .21 & 15.85 $\pm$ .28 \\ \midrule 
    + REX & \textcolor{myred}{\textbf{55.93}} $\pm$ .46 & \textcolor{myred}{\textbf{34.50}} $\pm$ .16 & \textcolor{myred}{\textbf{25.52}} $\pm$ .17 & \textbf{20.54} $\pm$ .32 & \textcolor{myred}{\textbf{16.97}} $\pm$ .46 & \textbf{14.60} $\pm$ .31 \\ \midrule
    Adam  & 58.65 $\pm$ 1.79 & 42.66 $\pm$ .68 & 33.17 $\pm$ 1.94 &  23.35 $\pm$ .20 & \textbf{19.63} $\pm$ .26 & 18.65 $\pm$ .07  \\
    + Step Schedule  & 59.35 $\pm$ .98 & 47.14 $\pm$ .42 & 35.10 $\pm$ 1.10 &  23.85 $\pm$ .07 & \textbf{19.63} $\pm$ .33 & \textbf{18.29} $\pm$ .10 \\  
    + Cosine Schedule  & 58.95 $\pm$ .95 & 40.69 $\pm$ 1.09 & \textbf{31.00} $\pm$ .74 &  22.85 $\pm$ .47 & 21.47 $\pm$ .31 & 19.08 $\pm$ .36\\    
    + OneCycle  & \textbf{57.88} $\pm$ .88 & {\textbf{36.41}} $\pm$ .29 & {\textbf{27.90}} $\pm$ .63 &  \textcolor{myred}{\textbf{20.02}} $\pm$ .19 & \textbf{19.21} $\pm$ .28 & 19.03 $\pm$ .43 \\  
    + Linear Schedule  & {\textbf{56.72}} $\pm$ .22 & \textbf{40.25} $\pm$ 1.00 & 31.15 $\pm$ .29 &  \textbf{21.70} $\pm$ .11 & 21.53 $\pm$ .44 & \textbf{17.85} $\pm$ .15  \\ 
    + Decay on Plateau   & 58.72 $\pm$ .60 & 42.30 $\pm$ .68 & 33.00 $\pm$ .80 &  22.77 $\pm$ .33 & 19.91 $\pm$ .45 & 19.61 $\pm$ .56 \\ 
    + Exp decay  & 58.92 $\pm$ .52 & 44.76 $\pm$ .90 & 33.52 $\pm$ 1.18 &  23.30 $\pm$ .39 & 20.70 $\pm$ .50 & 19.63 $\pm$ .24 \\ \midrule
    + REX & \textcolor{myred}{\textbf{56.47}} $\pm$ .31 & \textcolor{myred}{\textbf{35.52}} $\pm$ .44 & \textcolor{myred}{\textbf{27.24}} $\pm$ .20 & \textbf{21.65} $\pm$ .21 & \textcolor{myred}{\textbf{19.12}} $\pm$ .31 & \textcolor{myred}{\textbf{17.75}} $\pm$ .22 \\
 \bottomrule
\end{tabular}
\label{tablestl10}
\end{table*}
\begin{table*}[t]
\centering
\caption{\texttt{VGG16-CIFAR100} generalization error. The number of epochs was predefined before the execution of the algorithms. \textcolor{myred}{\textbf{Bold red}} indicates Top-1 performance, \textbf{black bold} is Top-3.}
\begin{tabular}{c|cccccc}
\toprule
    SGDM & 1\% & 5\% & 10\% & 25\% & 50\% & 100\% \\ \midrule
    + Step Schedule & 95.03 $\pm$ .42 & 69.87 $\pm$ .28 &  46.97 $\pm$ .13 & 35.04 $\pm$ .24 & 30.09 $\pm$ .32 & \textbf{27.83} $\pm$ .30 \\
    + Cosine Schedule & 95.03 $\pm$ .42 & 61.82 $\pm$ .13 & \textbf{41.26} $\pm$ .26 & \textbf{31.93} $\pm$ .09 & \textbf{28.63} $\pm$ .11 & \textbf{27.84} $\pm$ .12 \\    
    + OneCycle & \textcolor{myred}{\textbf{91.96}} $\pm$ 1.01 & \textbf{58.35} $\pm$ .40 & 45.39 $\pm$ .73 & 32.62 $\pm$ .21  & 30.10 $\pm$ .34 & 29.09 $\pm$ .12 \\  
    + Linear Schedule & 96.11 $\pm$ 1.64 & \textbf{58.14} $\pm$ 1.19 & \textcolor{myred}{\textbf{39.66}} $\pm$ .61 & \textbf{31.95} $\pm$ .29 & \textbf{29.10} $\pm$ .34 & 28.26 $\pm$ .08 \\ 
    + Decay on Plateau & \textbf{94.70} $\pm$ 1.20 & 65.25 $\pm$ 1.72 & 50.81 $\pm$ .58 & 35.29 $\pm$ .59 & 30.65 $\pm$ .31 & 29.74 $\pm$ .43 \\ 
    + Exp decay & 96.54 $\pm$ .39 & 65.65 $\pm$ 1.24 & 49.04 $\pm$ 1.98 & 33.15 $\pm$ .19 & 29.51 $\pm$ .22  & 28.47 $\pm$ .18 \\ \midrule
    + REX & \textbf{94.92} $\pm$ .91 & \textcolor{myred}{\textbf{56.62}} $\pm$ .65 & \textbf{40.72} $\pm$ .29 &  \textcolor{myred}{\textbf{31.16}} $\pm$ .11 & \textcolor{myred}{\textbf{28.54}} $\pm$ .02 &  \textcolor{myred}{\textbf{27.27}} $\pm$ .30 \\ \midrule
    Adam & 92.70 $\pm$ .50 & 64.05 $\pm$ .41 & 57.56 $\pm$ 1.30 & 37.98 $\pm$ .20 & 33.62 $\pm$ .11 & 31.09 $\pm$ .09 \\
    + Step Schedule & 92.65 $\pm$ .38 & 62.90 $\pm$ .08& 44.94 $\pm$ .49 & 34.16 $\pm$ .11 & 29.40 $\pm$ .22 & \textbf{27.75} $\pm$ .15 \\  
    + Cosine Schedule & \textcolor{myred}{\textbf{91.48}} $\pm$ .42 & \textbf{55.90} $\pm$ 2.46 & \textbf{40.31} $\pm$ .07 & \textbf{32.32} $\pm$ .14 & 29.68 $\pm$ .17 & \textbf{28.08}  $\pm$ .10 \\    
    + OneCycle & {\textbf{92.18}} $\pm$ .69 & 58.29 $\pm$ .53 & 43.47 $\pm$ .28 & 34.59 $\pm$ .31 & 29.83 $\pm$ .29 & 29.58 $\pm$ .18  \\  
    + Linear Schedule& 92.94 $\pm$ .49 & \textbf{54.32} $\pm$ 1.17 & \textcolor{myred}{\textbf{39.49}} $\pm$ .11 & \textbf{32.01} $\pm$ .49 & \textbf{29.30} $\pm$ .18 & 28.65 $\pm$ .10 \\ 
    + Decay on Plateau & 92.76 $\pm$ .48 & 64.10 $\pm$ .22 &  57.05 $\pm$ .84 & 32.60 $\pm$ .31 & \textbf{29.03} $\pm$ .10 & 28.67 $\pm$ .19 \\ 
    + Exp decay & 92.43 $\pm$ .67 & 55.26 $\pm$ 1.24 & 42.62 $\pm$ .12 & 32.37 $\pm$ .18 & 29.53 $\pm$ .12 & 28.83 $\pm$ .08  \\  \midrule
    + REX & \textbf{91.93} $\pm$ .01 & \textcolor{myred}{\textbf{52.20}} $\pm$ .47 & \textbf{39.51} $\pm$ .21 & \textcolor{myred}{\textbf{31.68}} $\pm$ .57 & \textcolor{myred}{\textbf{28.58}} $\pm$ .16 & \textcolor{myred}{\textbf{26.99}} $\pm$ .09 \\
 \bottomrule
\end{tabular}
\label{tablevgg}
\end{table*}

\subsection{Learning Rate Schedules}
There are many popular learning rate schedules implemented in widely-used frameworks and packages.
In general, the schedules are aware of the current time step $t$ and the maximum time step $T$. 
Let $\eta$ denote the learning rate and $\beta$ the momentum. 
We comprehensively detail the schedules considered in this paper below, covering almost all widely-implemented schedules; see Figure \ref{fig:schedulesAndRate} for a visualization.

\begin{itemize}[leftmargin=*]
    \item Step schedule \cite{he2016deep}: $\eta_t = \gamma_t \cdot \eta_0$ where $\gamma_t$ is piece-wise and depends on $\sfrac{t}{T}$. 
    A typical schedule \cite{he2016deep} 
    would be to decay the learning rate by $0.1$ at $\sfrac{1}{2}$ epochs and again by $0.1$ at $\sfrac{3}{4}$ epochs.
    We employ such a step schedule for all our experiments.
    \item Decay on Plateau \cite{tensorflow2015-whitepaper, paszke2017automatic}: A practical version of the step schedule, where the learning rate is decayed when the validation loss does not improve for certain number of tuneable epochs, which we tune in multiples of 5. 
    \item Linear schedule \cite{tensorflow2015-whitepaper, paszke2017automatic}: $\eta_t = \left(1 - \sfrac{t}{T}\right) \cdot \eta_0$.  
    \item Cosine schedule \cite{loshchilov2017sgdr}: $\eta_t = \tfrac{\eta_0}{2} \cdot \left(1 + \cos(\tfrac{\pi \cdot t}{T})\right)$.  
    \item Exponential schedule \cite{tensorflow2015-whitepaper, paszke2017automatic}: $\eta_t = \eta_0 \cdot e^{ \sfrac{\gamma t}{T}}$. We find that setting $\gamma=-3$ yields the best performance.  
    \item OneCycle schedule \cite{smith20181cycle}: \begin{equation*}
    \eta_t = \begin{cases}
    \eta_{\min} + (\eta_{\max} - \eta_{\min}) \cdot \left(\tfrac{t}{\sfrac{T}{2}}\right) \cdot \eta_0, & \textbf{if}~~ \sfrac{t}{T} < 1/2\\
    \eta_{\min} + (\eta_{\max} - \eta_{\min}) \cdot \left(2 - \tfrac{t}{\sfrac{T}{2}}\right) \cdot \eta_0, & \text{otherwise}
    \end{cases} \end{equation*}
    \begin{equation*}
    \beta_t = \begin{cases}
    \beta_{\min} + (\beta_{\max} - \beta_{\min}) \cdot \left(1 - \tfrac{t}{\sfrac{T}{2}}\right) \cdot \beta_0, & \textbf{if}~~ \sfrac{t}{T} < 1/2\\
    \beta_{\min} + (\beta_{\max} - \beta_{\min}) \cdot \left(\tfrac{t}{\sfrac{T}{2}} - 1\right) \cdot \beta_0, & \text{otherwise}
    \end{cases} \end{equation*}
    $\eta_{\min}$, $\eta_{\max}$, $\beta_{\min}$, and $\beta_{\max}$ are hyperparameters. 
    For fair computational comparison, we follow the recommended settings \cite{smith20181cycle} and set $\eta_{\min} = \eta_{\max} \cdot 0.1$, $\beta_{\max} = 0.95$, $\beta_{\min} = 0.85$, so that $\eta_{\max}$ is the only hyperparameter.
    \item REX schedule: 
    $$\eta_t = \eta_0 \cdot \left(\frac{1 - \sfrac{t}{T}}{ \sfrac{1}{2} + \sfrac{1}{2} \cdot (1 - \sfrac{t}{T})}\right).$$ 
    We re-emphasize the motivation for REX: it is a new profile and sampling rate combination, which is motivated by the improved performance of a delayed linear schedule in certain circumstances. REX aggressively decreases the learning rate towards the end of the training process, which is the ``reflection'' of the exponential decay. 
\end{itemize}

There are simply too many schedules to compare comprehensively, so we select the widely-used schedules above for comparison.
We apply the schedules to the two most popular optimizers: SGD with momentum and Adam.

\subsection{Empirical Results}

\begin{table*}[t]
\centering
\caption{\texttt{VAE-MNIST} generalization loss. The number of epochs was predefined before the execution of the algorithms. \textcolor{myred}{\textbf{Bold red}} indicates Top-1 performance, \textbf{black bold} is Top-3, ignoring non SGDM and Adam optimizers.}
\begin{tabular}{c|cccccc}
\toprule
    SGDM & 1\% & 5\% & 10\% & 25\% & 50\% & 100\% \\ \midrule
    + Step Schedule & 180.30 $\pm$ 6.98 & 152.97 $\pm$ .55 & 146.24 $\pm$ 2.50 &  140.28 $\pm$ .51 & 137.70 $\pm$ .93 & 136.34 $\pm$ .31\\
    + Cosine Schedule & 174.52 $\pm$ 1.09 & \textbf{145.99} $\pm$ .15 & \textbf{141.23} $\pm$ .36 &  \textbf{139.15} $\pm$ .26  & \textbf{136.69} $\pm$ .27 & \textbf{135.05} $\pm$ .09\\    
    + OneCycle  & \textbf{161.95} $\pm$ .67 & 146.25 $\pm$ .35 & \textbf{143.01} $\pm$ 1.08 &  \textbf{139.79} $\pm$ .66 & \textbf{137.20} $\pm$ .06 & \textbf{135.65} $\pm$ .44 \\  
    + Linear Schedule & 174.64 $\pm$ .15 & \textbf{146.15} $\pm$ .26 & 143.64 $\pm$ .80 &  148.00 $\pm$ .48 & 141.72 $\pm$ .48  & 137.84 $\pm$ .32 \\ 
    + Decay on Plateau & \textbf{167.16} $\pm$ .30 & 151.15 $\pm$ .11 & 146.82 $\pm$ .58 &  140.51 $\pm$ .73 & 139.54 $\pm$ .34 & 137.33 $\pm$ .49\\ 
    + Exp decay  & 179.60 $\pm$ 3.47 & 160.52 $\pm$ .64 & 146.24 $\pm$ .73 &  154.31 $\pm$ .43 & 145.83 $\pm$ .48 & 139.67 $\pm$ .57\\ \midrule
    + REX & \textcolor{myred}{\textbf{149.85}} $\pm$ 1.62 & \textcolor{myred}{\textbf{139.56}} $\pm$ .78 & \textcolor{myred}{\textbf{137.15}} $\pm$ .05   &  \textcolor{myred}{\textbf{134.41}} $\pm$ .78  & \textcolor{myred}{\textbf{135.69}} $\pm$ .24 & \textcolor{myred}{\textbf{135.03}} $\pm$ .37  \\ \midrule
    Adam & 152.10 $\pm$ .55 & 142.54 $\pm$ .50 & 140.10 $\pm$ .82 &  136.28 $\pm$ .18 & 134.64 $\pm$ .14 & 134.66 $\pm$ .17\\
    + Step Schedule & 153.45 $\pm$ 1.47 & 142.19 $\pm$ .98 & 138.32 $\pm$ .20 &  136.62 $\pm$ .30 & 134.14 $\pm$ .56 & 133.34 $\pm$ .41\\  
    + Cosine Schedule & 149.82 $\pm$ .32 & 140.78 $\pm$ .72 & \textbf{137.66} $\pm$ .79 &  134.73 $\pm$ .04 & \textbf{133.25} $\pm$ .26 & 133.23 $\pm$ .30\\    
    + OneCycle  & \textbf{149.07} $\pm$ .99 & \textbf{139.75} $\pm$ .27  & 138.12 $\pm$ .99 &  \textbf{134.67} $\pm$ .55 & \textbf{133.27} $\pm$ .07 & \textbf{132.83} $\pm$ .33\\  
    + Linear Schedule   & \textbf{148.93} $\pm$ .20 & \textbf{139.82} $\pm$ .20 & \textbf{137.00} $\pm$ .70 &  \textbf{134.71} $\pm$ .25 & 134.00 $\pm$ .49 & \textbf{132.95} $\pm$ .24\\ 
    + Decay on Plateau  & 152.08 $\pm$ .45 & 141.54 $\pm$ .31 & 139.76 $\pm$ .52 &  135.68 $\pm$ .59 & 134.10 $\pm$ .21 & 134.06 $\pm$ .45 \\ 
    + Exp decay & 149.28 $\pm$ .46 & 142.94 $\pm$ 1.28 & 138.82 $\pm$ .36 &  135.19 $\pm$ .43 & 134.05 $\pm$ .16 & 133.88 $\pm$ .85\\  \midrule
    + REX & \textcolor{myred}{\textbf{148.59}} $\pm$ .33 & \textcolor{myred}{\textbf{139.05}} $\pm$ .20 & \textcolor{myred}{\textbf{136.62}} $\pm$ .21 & \textcolor{myred}{\textbf{134.24}} $\pm$ .02 & \textcolor{myred}{\textbf{133.16}} $\pm$ .05 & \textcolor{myred}{\textbf{132.52}} $\pm$ .05 \\
 \bottomrule
\end{tabular}
\label{tablevae}
\end{table*}

\textbf{Image Classification.} 
We choose four diverse settings for this task. 
For datasets, we use the standard CIFAR10 and CIFAR100 datasets, in addition to the low count, high-res STL10 dataset, as well as the standard ImageNet dataset. 
Since ResNets remain the most commonly-deployed model in industry, we perform experiments with three variations of the ResNet \cite{he2016deep}. 
The ResNet20 comes from the line of lower cost, lower performance ResNets, and is a close cousin of the more expensive and better performing ResNet18. 
ResNet50 belongs to the latter series, and is a standard model for ImageNet. 
We also include the Wide ResNet variation which further increases the model width for better performance \cite{zagoruyko2016wide}. 
The other model we employ is the VGG-16 model \cite{simonyan2014very}. 
While VGG models are far outdated in attaining state-of-the-art performance, the architecture is still relevant for custom applications with smaller CNNs, where residual connections have limited application. 
We provide thorough evaluation in the \texttt{RN20-CIFAR10}, \texttt{WRN-STL10}, \texttt{VGG16-CIFAR100} settings, and, due to computational constraints, provide lower epochs results for \texttt{RN50-ImageNet}, given in Tables \ref{tablern20cifar10}, \ref{tablestl10}, \ref{tablevgg}, and \ref{tableimagenet}.

As observed in \cite{li2020budgeted}, the linear schedule performs well for both SGD and Adam, particularly for a low number of epochs. 
While the Step schedule performs well for the maximum number of epochs, it scales very poorly to lower epoch settings. 
On the other hand, REX performs well in both high and low epoch regimes. 
Results also follow general Computer Vision observations for these settings, where SGD tends to outperform Adam.

\begin{table}[!ht]
\centering
\caption{\texttt{RN50-ImageNet} generalization error. The number of epochs was predefined before the execution of the algorithms. \textcolor{myred}{\textbf{Bold red}} indicates Top-1 performance, \textbf{black bold} is Top-3.}
\begin{tabular}{c|cc}
\toprule
    SGDM & 1\% & 5\% \\ \midrule
    + Step Schedule & 87.28 & 46.58 \\
    + Cosine Schedule & \textbf{82.88} & \textbf{43.90} \\    
    + OneCycle & 90.94 & 55.00 \\  
    + Linear Schedule & \textbf{82.00} & \textbf{43.27}\\ 
    + Exp decay & 90.19 & 48.28\\ \midrule
    + REX & \textcolor{myred}{\textbf{80.98}} & \textcolor{myred}{\textbf{40.78}}\\ \midrule
    Adam & 1\% & 5\%\\ \midrule
    + Step Schedule & 77.97 & 45.91\\
    + Cosine Schedule & \textbf{73.51} & \textbf{43.66} \\    
    + OneCycle & 82.58 & 62.57 \\  
    + Linear Schedule & \textbf{71.42} & \textbf{42.01} \\ 
    + Exp decay & 75.54 & 45.43 \\ \midrule
    + REX & \textcolor{myred}{\textbf{69.91}} & \textcolor{myred}{\textbf{40.65}}\\ \midrule
 \bottomrule
\end{tabular}
\label{tableimagenet}
\end{table}

\begin{table*}[t]
\centering
\caption{\texttt{YOLO-VOC} mAP. The number of epochs was predefined before the execution of the algorithms. \textcolor{myred}{\textbf{Bold red}} indicates Top-1 performance, \textbf{black bold} is Top-3.}
\begin{tabular}{c|cccccc}
\toprule
     & 1\% & 5\% & 10\% & 25\% & 50\% & 100\% \\ \midrule
    Adam & 45.0 $\pm$ 3.4 & 48.1 $\pm$ 7.6 & 61.9 $\pm$ 1.8 & 70.2 $\pm$ 3.5 & 72.1 $\pm$ 6.4  & 79.1 $\pm$ 1.6 \\
    + Step Schedule & 62.2 $\pm$ 1.7 & \textbf{67.0} $\pm$ 3.4 & 71.8 $\pm$ 1.0 & 78.5 $\pm$ 0.2 & 81.1 $\pm$ 1.0 & 83.2 $\pm$ 0.2 \\
    + OneCycle & 60.4 $\pm$ 7.2 & 63.8 $\pm$ 7.6 & 74.9 $\pm$ 1.0 & 79.9 $\pm$ 1.3 & 81.1 $\pm$ 2.8 & 83.3 $\pm$ 0.4 \\
    + Cosine Schedule & \textbf{63.6} $\pm$ 5.2 & 66.8 $\pm$ 6.1 & \textbf{75.9} $\pm$ 0.2 & \textbf{81.1} $\pm$ 0.7  & \textcolor{myred}{\textbf{82.5}} $\pm$ 1.0 & \textcolor{myred}{\textbf{84.0}} $\pm$ 0.2 \\
    + Linear Schedule & \textbf{63.7} $\pm$ 5.5 & {\textbf{67.2}} $\pm$ 5.9 & \textbf{76.2} $\pm$ 0.7 & \textbf{81.1} $\pm$ 0.9 & \textbf{82.4} $\pm$ 1.2 & \textbf{83.4} $\pm$ 0.2 \\ 
    + Exp decay& 49.6 $\pm$ 24 & \textcolor{myred}{\textbf{68.1}} $\pm$ 4.6 & 75.6 $\pm$ 0.1 & 80.1 $\pm$ 0.7 & 81.2 $\pm$ 2.2 & 83.2 $\pm$ 0.2 \\\midrule
    + REX & \textcolor{myred}{\textbf{64.0}} $\pm$ 5.0 & \textbf{67.0} $\pm$ 6.5 & \textcolor{myred}{\textbf{76.7}} $\pm$ 0.3 & \textcolor{myred}{\textbf{81.2}} $\pm$ 0.7 & \textbf{82.2} $\pm$ 1.8 & \textbf{83.4} $\pm$ 0.4 \\
 \bottomrule
\end{tabular}
\label{tablevoc}
\end{table*}

\textbf{Image Generation.} 
The two most popular types of networks for image generation are Variational Encoders (VAE) \cite{kingma2015vae} and Generative Adversarial Networks (GAN) \cite{goodfellow2014generative}. 
However, out of the two, only VAEs consistently benefit from learning rate decay \cite{goodfellow2014generative, chen2016infogan, arjovsky2017wasserstein, brock2019large, hou2016deep, sonderby2016ladder, vahdat2021nvae}. 
Therefore, we select VAEs as the network of choice for image generation.
We train VAEs on the MNIST dataset for 200 epochs, after which performance no longer improves. 
Results are given in Table \ref{tablevae}. 

The linear schedule performs well for Adam, but not for SGDM. 
Similarly, the cosine schedule performs well for SGDM, but not for Adam. 
The OneCycle schedule performs well across all settings, but REX outperforms all other schedules in the low budget and high budget setting. 

\textbf{Object Detection.} 
We train a YOLOv3 \cite{redmon2018yolov3} model on the Pascal VOC dataset. 
The training set is the combined 2007 and 2012 training set, and the test set is the 2007 test set. 
We were able to achieve the mAP score reported in the literature by training the network for 50 epochs.
Thus, we set this as the maximum number of epochs. 
We find that the network does not train well without a warm-up period, so all networks are trained for 2 epochs from a learning rate of 1e-5 linearly increased to 1e-4. 
This warm-up phase is not counted as part of the allocated training budget.
We also round up the number of epochs to the closest integer: for example, the 1\% setting trains for 2 warmup epochs and then $\lceil 50 \cdot  0.01 \rceil = 1$ epoch, for a total of 3 epochs. The 100\% setting trains for 2 warmup epochs and then 50 epochs for a total of 52 epochs. 
Results are given in Table \ref{tablevoc}.
Similar to other settings, the step schedule performs reasonably well for a large number of epochs, but is outperformed by the cosine schedule. 
REX performs well in the low epoch setting.

\textbf{Natural Language Processing.} 
Fine-tuning pre-trained transformer models is one of the most common training procedures in NLP \cite{devlin2019bert, brown2020language}, thus making it a setting of interest.
This is because $i)$ it is often cost-prohibitive for practitioners to pre-train their own models and $ii)$ fine-tuning pre-trained transformers often results in significantly better performance in comparison to training a smaller model from scratch. 
The linear schedule is the default schedule implemented in HuggingFace \cite{wolf2020huggingfaces}, the most popular package for transformer models, and is considered the gold standard in this domain.
We fine-tune \texttt{BERT$_{BASE}$} on the GLUE benchmark, an NLP benchmark with nine datasets. 
We leave out the problematic WNLI dataset \cite{devlin2019bert}. 
Since we are able to attain the scores reported in the literature with 3 epochs of fine-tuning, we set that as the maximum number of epochs. 
Due to computational constraints, we can only perform one run per setting, which causes some variability within the results.
Although REX achieves the best mean score for small and large budgets, we see that the best optimizer can vary depending on the dataset. 
For example, OneCycle attains the best scores on QNLI and MRPC, and the Cosine schedule performs the best on SST-2.

\begin{table}
\centering
\caption{Results of \texttt{BERT$_{BASE}$-GLUE}. AdamW + Linear Schedule follows the huggingface \cite{wolf2020huggingfaces} implementation, and achieves the results in well-known studies \cite{devlin2019bert, sanh2020distilbert}. Results given by 1 epoch/2 epochs/3 epochs. Excluding the problematic WNLI dataset \cite{devlin2019bert}. }
\begin{tabular}{c|c} 
\toprule
    & Score\\ \midrule 
    AdamW & 79.9/81.2/81.8 \\
    + Step Schedule & 80.2/81.9/82.3 \\
    + Cosine Schedule & 80.9/\textbf{82.2}/\textbf{82.7}  \\ 
    + OneCycle & \textbf{81.0}/82.0/\textbf{82.7} \\ 
    + Linear Schedule & \textbf{81.2}/\textbf{82.3}/82.6  \\ 
    + Exp decay & 80.6/81.8/82.5\\ \midrule 
    + REX & \textcolor{myred}{\textbf{81.7}}/\textcolor{myred}{\textbf{82.6}}/\textcolor{myred}{\textbf{82.8}} \\
 \bottomrule
\end{tabular}
\label{tablebertgluecompsummary}
\end{table}

\begin{table*}
\centering
\begin{footnotesize}
\caption{\small Results of \texttt{BERT$_{BASE}$-GLUE}. AdamW + Linear Schedule follows the huggingface \cite{wolf2020huggingfaces} implementation, and achieves the results in well-known studies \cite{devlin2019bert, sanh2020distilbert}. Results given by 1 epoch/2 epochs/3 epochs. Excluding the problematic WNLI dataset \cite{devlin2019bert}. 
}
\begin{tabular}{c|ccccccccc} 
\toprule
    & CoLA & MNLI & MRPC & QNLI  & QQP & RTE & SST-2 & STS-B & \\ \midrule 
    AdamW & 54.8/54.7/55.2 & 82.9/83.3/83.7 & 84.8/87.2/87.6 & 88.7/\textbf{90.4}/\textcolor{myred}{\textbf{90.7}} & 89.4/90.2/90.5 & 59.2/64.6/66.8 & 91.2/91.3/91.2 & \textbf{87.8}/87.8/88.3 \\
    + Step Schedule & 53.5/56.9/56.6 & 82.6/83.4/83.9 & 85.6/87.9/88.3 & 88.2/90.1/\textbf{90.4} & 89.0/90.5/90.6 & \textbf{63.5}/\textbf{65.7}/\textcolor{myred}{\textbf{67.5}} & \textcolor{myred}{\textbf{92.8}}/92.8/\textbf{93.0}  & 86.7/88.0/88.4 \\
    + Cosine Schedule & 55.7/\textbf{58.6}/58.2 & \textbf{83.5}/\textbf{84.0}/\textbf{84.2} & 84.5/87.6/87.9 & \textbf{89.4}/89.8/\textbf{90.4} & \textbf{89.8}/\textbf{90.6}/\textcolor{myred}{\textbf{91.0}} & \textbf{64.2}/65.3/\textcolor{myred}{\textbf{67.5}} & \textbf{92.7}/\textcolor{myred}{\textbf{93.1}}/\textcolor{myred}{\textbf{93.7}} & 87.4/88.4/\textbf{88.7} \\ 
    + OneCycle & \textbf{57.7}/\textbf{58.1}/56.5 & \textcolor{myred}{\textbf{83.6}}/83.8/\textbf{84.2} & \textcolor{myred}{\textbf{87.3}}/87.5/\textcolor{myred}{\textbf{89.9}} & \textcolor{myred}{\textbf{89.5}}/\textcolor{myred}{\textbf{91.0}}/\textcolor{myred}{\textbf{90.7}} & \textbf{89.8}/\textbf{90.6}/90.8 & 60.3/63.9/\textcolor{myred}{\textbf{67.5}} & 92.1/92.2/\textbf{93.0} & \textcolor{myred}{\textbf{88.1}}/{\textbf{88.5}}/\textcolor{myred}{\textbf{89.0}} \\ 
    + Linear Schedule  & \textcolor{myred}{\textbf{58.0}}/57.6/\textbf{58.8} & \textbf{83.5}/\textcolor{myred}{\textbf{84.1}}/\textcolor{myred}{\textbf{84.3}} & 85.4/\textbf{88.1}/88.0 & 88.8/\textbf{90.4}/89.6 & 89.7/\textbf{90.6}/\textcolor{myred}{\textbf{91.0}} & \textbf{63.5}/\textbf{65.7}/67.1 & \textcolor{myred}{\textbf{92.8}}/\textbf{93.0}/92.9 & \textbf{87.9}/{\textbf{88.5}}/\textbf{88.8} \\ 
    + Exp decay & 57.5/57.3/\textcolor{myred}{\textbf{59.1}} & \textcolor{myred}{\textbf{83.6}}/83.9/84.1 & \textbf{86.2}/\textbf{88.7}/\textbf{89.1} & 88.2/89.2/89.6 & 88.8/90.3/90.6 & 61.0/63.9/66.0 & 92.1/\textcolor{myred}{\textbf{93.1}}/\textbf{93.0} & 87.2/88.2/88.5 \\ \midrule 
    + REX & \textbf{57.8}/\textcolor{myred}{\textbf{58.8}}/\textcolor{myred}{\textbf{59.1}} & 83.4/\textbf{84.0}/\textcolor{myred}{\textbf{84.3}} & \textcolor{myred}{\textbf{87.3}}/\textcolor{myred}{\textbf{88.9}}/\textbf{89.1} & \textbf{88.9}/\textbf{90.5}/90.3 & \textcolor{myred}{\textbf{90.0}}/\textcolor{myred}{\textbf{90.7}}/\textcolor{myred}{\textbf{91.0}} & \textcolor{myred}{\textbf{65.3}}/\textcolor{myred}{\textbf{66.8}}/67.1 & \textbf{92.7}/92.7/92.7 & 87.6/\textcolor{myred}{\textbf{88.6}}/88.6 \\
 \bottomrule
\end{tabular}
\label{tablebertgluecomp}
\end{footnotesize}
\end{table*}

\begin{figure*}
  \centering
    \begin{subfigure}[b]{0.24\linewidth}
    \includegraphics[width=\linewidth]{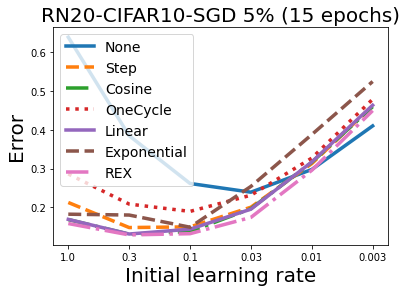}
    \label{fig:RN20CIFAR105EPOCHSHyper}
  \end{subfigure}
  \begin{subfigure}[b]{0.24\linewidth}
    \includegraphics[width=\linewidth]{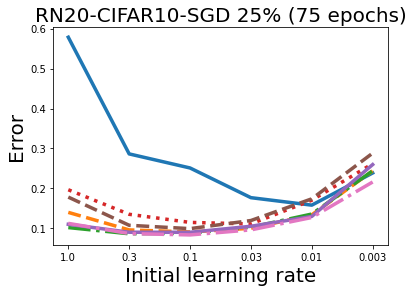}
    \label{fig:RN20CIFAR1075EPOCHSHyper}
  \end{subfigure} 
  \begin{subfigure}[b]{0.24\linewidth}
    \includegraphics[width=\linewidth]{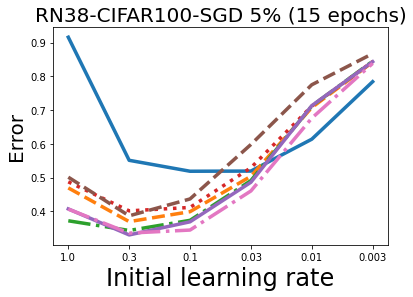}
    \label{fig:RN38CIFAR1005EPOCHSHyper}
  \end{subfigure} 
  \begin{subfigure}[b]{0.24\linewidth}
    \includegraphics[width=\linewidth]{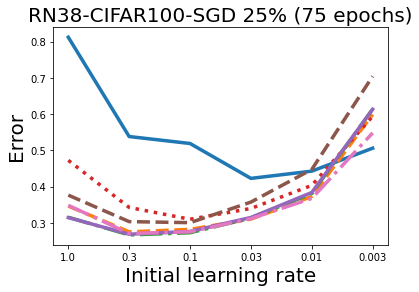}
    \label{fig:RN38CIFAR10075EPOCHSHyper}
  \end{subfigure}
  \caption{\small Error against initial learning for \texttt{RN20-CIFAR10-SGD} and \texttt{RN38-CIFAR100-SGD} for 5\% and 25\% of total epochs. As expected all, schedules suffer as the learning rate grows too large or too small.}
  \label{fig:hyperparameterSensitivity}
\end{figure*}

\textbf{Sensitivity to learning rate tuning.} While it is reasonable to suggest that the practitioner simply pick a per-iteration sampling rate for the REX, linear, and other profiles, a relevant issue in budgeted training is performance given a limited number of experimental trials.
Namely, in extreme cases, the practitioner may not even have the budget to finely tune the learning rate.
Therefore, we plot the considered schedules in two settings against learning rate, presented in Figure \ref{fig:hyperparameterSensitivity}.
Clearly, there is no schedule that can recover from a poor initial learning rate.
However, schedules tend to retain their relative ordering across initial learning rates.
This means that even with poor hyperparameter settings, the choice of learning rate schedule remains important.
REX, represented by the pink line below all other lines, outperforms other schedules for most learning rates in the budgeted settings presented in the plots.



\section{Conclusion}
In this paper, we identified issues with existing learning rate schedules in the budgeted setting. We proposed a profile and sampling rate framework for understanding existing schedules. While there is no optimal profile, we found that the proposed REX schedule performs well with a sampling rate of every iteration in both small and large epoch regimes. With thorough empirical evaluation, we confirm that the proposed REX learning rate schedule performs favorably across a large number of settings including image classification, generation, object detection, and natural language processing.


\bibliographystyle{ACM-Reference-Format}
\bibliography{iclr2021_conference}


\begin{thebibliography}{50}


\ifx \showCODEN    \undefined \def \showCODEN     #1{\unskip}     \fi
\ifx \showDOI      \undefined \def \showDOI       #1{#1}\fi
\ifx \showISBNx    \undefined \def \showISBNx     #1{\unskip}     \fi
\ifx \showISBNxiii \undefined \def \showISBNxiii  #1{\unskip}     \fi
\ifx \showISSN     \undefined \def \showISSN      #1{\unskip}     \fi
\ifx \showLCCN     \undefined \def \showLCCN      #1{\unskip}     \fi
\ifx \shownote     \undefined \def \shownote      #1{#1}          \fi
\ifx \showarticletitle \undefined \def \showarticletitle #1{#1}   \fi
\ifx \showURL      \undefined \def \showURL       {\relax}        \fi
\providecommand\bibfield[2]{#2}
\providecommand\bibinfo[2]{#2}
\providecommand\natexlab[1]{#1}
\providecommand\showeprint[2][]{arXiv:#2}

\bibitem[\protect\citeauthoryear{Abadi, Agarwal, Barham, Brevdo, Chen, Citro,
  Corrado, Davis, Dean, Devin, Ghemawat, Goodfellow, Harp, Irving, Isard, Jia,
  Jozefowicz, Kaiser, Kudlur, Levenberg, Man\'{e}, Monga, Moore, Murray, Olah,
  Schuster, Shlens, Steiner, Sutskever, Talwar, Tucker, Vanhoucke, Vasudevan,
  Vi\'{e}gas, Vinyals, Warden, Wattenberg, Wicke, Yu, and Zheng}{Abadi
  et~al\mbox{.}}{2015}]%
        {tensorflow2015-whitepaper}
\bibfield{author}{\bibinfo{person}{Mart\'{\i}n Abadi}, \bibinfo{person}{Ashish
  Agarwal}, \bibinfo{person}{Paul Barham}, \bibinfo{person}{Eugene Brevdo},
  \bibinfo{person}{Zhifeng Chen}, \bibinfo{person}{Craig Citro},
  \bibinfo{person}{Greg~S. Corrado}, \bibinfo{person}{Andy Davis},
  \bibinfo{person}{Jeffrey Dean}, \bibinfo{person}{Matthieu Devin},
  \bibinfo{person}{Sanjay Ghemawat}, \bibinfo{person}{Ian Goodfellow},
  \bibinfo{person}{Andrew Harp}, \bibinfo{person}{Geoffrey Irving},
  \bibinfo{person}{Michael Isard}, \bibinfo{person}{Yangqing Jia},
  \bibinfo{person}{Rafal Jozefowicz}, \bibinfo{person}{Lukasz Kaiser},
  \bibinfo{person}{Manjunath Kudlur}, \bibinfo{person}{Josh Levenberg},
  \bibinfo{person}{Dan Man\'{e}}, \bibinfo{person}{Rajat Monga},
  \bibinfo{person}{Sherry Moore}, \bibinfo{person}{Derek Murray},
  \bibinfo{person}{Chris Olah}, \bibinfo{person}{Mike Schuster},
  \bibinfo{person}{Jonathon Shlens}, \bibinfo{person}{Benoit Steiner},
  \bibinfo{person}{Ilya Sutskever}, \bibinfo{person}{Kunal Talwar},
  \bibinfo{person}{Paul Tucker}, \bibinfo{person}{Vincent Vanhoucke},
  \bibinfo{person}{Vijay Vasudevan}, \bibinfo{person}{Fernanda Vi\'{e}gas},
  \bibinfo{person}{Oriol Vinyals}, \bibinfo{person}{Pete Warden},
  \bibinfo{person}{Martin Wattenberg}, \bibinfo{person}{Martin Wicke},
  \bibinfo{person}{Yuan Yu}, {and} \bibinfo{person}{Xiaoqiang Zheng}.}
  \bibinfo{year}{2015}\natexlab{}.
\newblock \bibinfo{title}{{TensorFlow}: Large-Scale Machine Learning on
  Heterogeneous Systems}.
\newblock
\newblock
\urldef\tempurl%
\url{http://tensorflow.org/}
\showURL{%
\tempurl}
\newblock
\shownote{Software available from tensorflow.org.}


\bibitem[\protect\citeauthoryear{Arjovsky, Chintala, and Bottou}{Arjovsky
  et~al\mbox{.}}{2017}]%
        {arjovsky2017wasserstein}
\bibfield{author}{\bibinfo{person}{Martin Arjovsky}, \bibinfo{person}{Soumith
  Chintala}, {and} \bibinfo{person}{Léon Bottou}.}
  \bibinfo{year}{2017}\natexlab{}.
\newblock \bibinfo{title}{Wasserstein GAN}.
\newblock
\newblock
\showeprint[arxiv]{1701.07875}~[stat.ML]


\bibitem[\protect\citeauthoryear{Brock, Donahue, and Simonyan}{Brock
  et~al\mbox{.}}{2019}]%
        {brock2019large}
\bibfield{author}{\bibinfo{person}{Andrew Brock}, \bibinfo{person}{Jeff
  Donahue}, {and} \bibinfo{person}{Karen Simonyan}.}
  \bibinfo{year}{2019}\natexlab{}.
\newblock \bibinfo{title}{Large Scale GAN Training for High Fidelity Natural
  Image Synthesis}.
\newblock
\newblock
\showeprint[arxiv]{1809.11096}~[cs.LG]


\bibitem[\protect\citeauthoryear{Brown, Mann, Ryder, Subbiah, Kaplan, Dhariwal,
  Neelakantan, Shyam, Sastry, Askell, Agarwal, Herbert-Voss, Krueger, Henighan,
  Child, Ramesh, Ziegler, Wu, Winter, Hesse, Chen, Sigler, Litwin, Gray, Chess,
  Clark, Berner, McCandlish, Radford, Sutskever, and Amodei}{Brown
  et~al\mbox{.}}{2020}]%
        {brown2020language}
\bibfield{author}{\bibinfo{person}{Tom~B. Brown}, \bibinfo{person}{Benjamin
  Mann}, \bibinfo{person}{Nick Ryder}, \bibinfo{person}{Melanie Subbiah},
  \bibinfo{person}{Jared Kaplan}, \bibinfo{person}{Prafulla Dhariwal},
  \bibinfo{person}{Arvind Neelakantan}, \bibinfo{person}{Pranav Shyam},
  \bibinfo{person}{Girish Sastry}, \bibinfo{person}{Amanda Askell},
  \bibinfo{person}{Sandhini Agarwal}, \bibinfo{person}{Ariel Herbert-Voss},
  \bibinfo{person}{Gretchen Krueger}, \bibinfo{person}{Tom Henighan},
  \bibinfo{person}{Rewon Child}, \bibinfo{person}{Aditya Ramesh},
  \bibinfo{person}{Daniel~M. Ziegler}, \bibinfo{person}{Jeffrey Wu},
  \bibinfo{person}{Clemens Winter}, \bibinfo{person}{Christopher Hesse},
  \bibinfo{person}{Mark Chen}, \bibinfo{person}{Eric Sigler},
  \bibinfo{person}{Mateusz Litwin}, \bibinfo{person}{Scott Gray},
  \bibinfo{person}{Benjamin Chess}, \bibinfo{person}{Jack Clark},
  \bibinfo{person}{Christopher Berner}, \bibinfo{person}{Sam McCandlish},
  \bibinfo{person}{Alec Radford}, \bibinfo{person}{Ilya Sutskever}, {and}
  \bibinfo{person}{Dario Amodei}.} \bibinfo{year}{2020}\natexlab{}.
\newblock \bibinfo{title}{Language Models are Few-Shot Learners}.
\newblock
\newblock
\showeprint[arxiv]{2005.14165}~[cs.CL]


\bibitem[\protect\citeauthoryear{Chang, Meng, Haber, Ruthotto, Begert, and
  Holtham}{Chang et~al\mbox{.}}{2017}]%
        {chang2017reversible}
\bibfield{author}{\bibinfo{person}{Bo Chang}, \bibinfo{person}{Lili Meng},
  \bibinfo{person}{Eldad Haber}, \bibinfo{person}{Lars Ruthotto},
  \bibinfo{person}{David Begert}, {and} \bibinfo{person}{Elliot Holtham}.}
  \bibinfo{year}{2017}\natexlab{}.
\newblock \bibinfo{title}{Reversible Architectures for Arbitrarily Deep
  Residual Neural Networks}.
\newblock
\newblock
\showeprint[arxiv]{1709.03698}~[cs.CV]


\bibitem[\protect\citeauthoryear{Chen, Wolfe, Li, and Kyrillidis}{Chen
  et~al\mbox{.}}{2020b}]%
        {chen2020demon}
\bibfield{author}{\bibinfo{person}{John Chen}, \bibinfo{person}{Cameron Wolfe},
  \bibinfo{person}{Zhao Li}, {and} \bibinfo{person}{Anastasios Kyrillidis}.}
  \bibinfo{year}{2020}\natexlab{b}.
\newblock \bibinfo{title}{Demon: Momentum Decay for Improved Neural Network
  Training}.
\newblock
\newblock
\showeprint[arxiv]{1910.04952}~[cs.LG]


\bibitem[\protect\citeauthoryear{Chen, Kornblith, Swersky, Norouzi, and
  Hinton}{Chen et~al\mbox{.}}{2020a}]%
        {chen2020big}
\bibfield{author}{\bibinfo{person}{Ting Chen}, \bibinfo{person}{Simon
  Kornblith}, \bibinfo{person}{Kevin Swersky}, \bibinfo{person}{Mohammad
  Norouzi}, {and} \bibinfo{person}{Geoffrey Hinton}.}
  \bibinfo{year}{2020}\natexlab{a}.
\newblock \bibinfo{title}{Big Self-Supervised Models are Strong Semi-Supervised
  Learners}.
\newblock
\newblock
\showeprint[arxiv]{2006.10029}~[cs.LG]


\bibitem[\protect\citeauthoryear{Chen, Duan, Houthooft, Schulman, Sutskever,
  and Abbeel}{Chen et~al\mbox{.}}{2016}]%
        {chen2016infogan}
\bibfield{author}{\bibinfo{person}{Xi Chen}, \bibinfo{person}{Yan Duan},
  \bibinfo{person}{Rein Houthooft}, \bibinfo{person}{John Schulman},
  \bibinfo{person}{Ilya Sutskever}, {and} \bibinfo{person}{Pieter Abbeel}.}
  \bibinfo{year}{2016}\natexlab{}.
\newblock \bibinfo{title}{InfoGAN: Interpretable Representation Learning by
  Information Maximizing Generative Adversarial Nets}.
\newblock
\newblock
\showeprint[arxiv]{1606.03657}~[cs.LG]


\bibitem[\protect\citeauthoryear{Devlin, Chang, Lee, and Toutanova}{Devlin
  et~al\mbox{.}}{2019}]%
        {devlin2019bert}
\bibfield{author}{\bibinfo{person}{Jacob Devlin}, \bibinfo{person}{Ming-Wei
  Chang}, \bibinfo{person}{Kenton Lee}, {and} \bibinfo{person}{Kristina
  Toutanova}.} \bibinfo{year}{2019}\natexlab{}.
\newblock \bibinfo{title}{BERT: Pre-training of Deep Bidirectional Transformers
  for Language Understanding}.
\newblock
\newblock
\showeprint[arxiv]{1810.04805}~[cs.CL]


\bibitem[\protect\citeauthoryear{Duchi, Hazan, and Singer}{Duchi
  et~al\mbox{.}}{2011}]%
        {duchi2011adaptive}
\bibfield{author}{\bibinfo{person}{John Duchi}, \bibinfo{person}{Elad Hazan},
  {and} \bibinfo{person}{Yoram Singer}.} \bibinfo{year}{2011}\natexlab{}.
\newblock \showarticletitle{Adaptive subgradient methods for online learning
  and stochastic optimization}.
\newblock \bibinfo{journal}{\emph{Journal of Machine Learning Research}}
  \bibinfo{volume}{12}, \bibinfo{number}{Jul} (\bibinfo{year}{2011}),
  \bibinfo{pages}{2121--2159}.
\newblock


\bibitem[\protect\citeauthoryear{Goodfellow, Pouget-Abadie, Mirza, Xu,
  Warde-Farley, Ozair, Courville, and Bengio}{Goodfellow et~al\mbox{.}}{2014}]%
        {goodfellow2014generative}
\bibfield{author}{\bibinfo{person}{Ian~J. Goodfellow}, \bibinfo{person}{Jean
  Pouget-Abadie}, \bibinfo{person}{Mehdi Mirza}, \bibinfo{person}{Bing Xu},
  \bibinfo{person}{David Warde-Farley}, \bibinfo{person}{Sherjil Ozair},
  \bibinfo{person}{Aaron Courville}, {and} \bibinfo{person}{Yoshua Bengio}.}
  \bibinfo{year}{2014}\natexlab{}.
\newblock \bibinfo{title}{Generative Adversarial Networks}.
\newblock
\newblock
\showeprint[arxiv]{1406.2661}~[stat.ML]


\bibitem[\protect\citeauthoryear{Goyal, Dollár, Girshick, Noordhuis,
  Wesolowski, Kyrola, Tulloch, Jia, and He}{Goyal et~al\mbox{.}}{2018}]%
        {goyal2018accurate}
\bibfield{author}{\bibinfo{person}{Priya Goyal}, \bibinfo{person}{Piotr
  Dollár}, \bibinfo{person}{Ross Girshick}, \bibinfo{person}{Pieter
  Noordhuis}, \bibinfo{person}{Lukasz Wesolowski}, \bibinfo{person}{Aapo
  Kyrola}, \bibinfo{person}{Andrew Tulloch}, \bibinfo{person}{Yangqing Jia},
  {and} \bibinfo{person}{Kaiming He}.} \bibinfo{year}{2018}\natexlab{}.
\newblock \bibinfo{title}{Accurate, Large Minibatch SGD: Training ImageNet in 1
  Hour}.
\newblock
\newblock
\showeprint[arxiv]{1706.02677}~[cs.CV]


\bibitem[\protect\citeauthoryear{He, Gkioxari, Dollár, and Girshick}{He
  et~al\mbox{.}}{2018}]%
        {he2018mask}
\bibfield{author}{\bibinfo{person}{Kaiming He}, \bibinfo{person}{Georgia
  Gkioxari}, \bibinfo{person}{Piotr Dollár}, {and} \bibinfo{person}{Ross
  Girshick}.} \bibinfo{year}{2018}\natexlab{}.
\newblock \bibinfo{title}{Mask R-CNN}.
\newblock
\newblock
\showeprint[arxiv]{1703.06870}~[cs.CV]


\bibitem[\protect\citeauthoryear{He, Zhang, Ren, and Sun}{He
  et~al\mbox{.}}{2016}]%
        {he2016deep}
\bibfield{author}{\bibinfo{person}{Kaiming He}, \bibinfo{person}{Xiangyu
  Zhang}, \bibinfo{person}{Shaoqing Ren}, {and} \bibinfo{person}{Jian Sun}.}
  \bibinfo{year}{2016}\natexlab{}.
\newblock \showarticletitle{Deep residual learning for image recognition}. In
  \bibinfo{booktitle}{\emph{Proceedings of the IEEE conference on computer
  vision and pattern recognition}}. \bibinfo{pages}{770--778}.
\newblock


\bibitem[\protect\citeauthoryear{Hinton, Srivastava, and Swersky}{Hinton
  et~al\mbox{.}}{2012}]%
        {hinton2012neural}
\bibfield{author}{\bibinfo{person}{Geoffrey Hinton}, \bibinfo{person}{Nitish
  Srivastava}, {and} \bibinfo{person}{Kevin Swersky}.}
  \bibinfo{year}{2012}\natexlab{}.
\newblock \showarticletitle{Neural networks for machine learning lecture 6a
  overview of mini-batch gradient descent}.
\newblock \bibinfo{journal}{\emph{Cited on}}  \bibinfo{volume}{14}
  (\bibinfo{year}{2012}), \bibinfo{pages}{8}.
\newblock


\bibitem[\protect\citeauthoryear{Hou, Shen, Sun, and Qiu}{Hou
  et~al\mbox{.}}{2016}]%
        {hou2016deep}
\bibfield{author}{\bibinfo{person}{Xianxu Hou}, \bibinfo{person}{Linlin Shen},
  \bibinfo{person}{Ke Sun}, {and} \bibinfo{person}{Guoping Qiu}.}
  \bibinfo{year}{2016}\natexlab{}.
\newblock \bibinfo{title}{Deep Feature Consistent Variational Autoencoder}.
\newblock
\newblock
\showeprint[arxiv]{1610.00291}~[cs.CV]


\bibitem[\protect\citeauthoryear{Hu, Shen, Albanie, Sun, and Wu}{Hu
  et~al\mbox{.}}{2017}]%
        {hu2017squeeze}
\bibfield{author}{\bibinfo{person}{Jie Hu}, \bibinfo{person}{Li Shen},
  \bibinfo{person}{Samuel Albanie}, \bibinfo{person}{Gang Sun}, {and}
  \bibinfo{person}{Enhua Wu}.} \bibinfo{year}{2017}\natexlab{}.
\newblock \showarticletitle{Squeeze-and-excitation networks}.
\newblock \bibinfo{journal}{\emph{arxiv preprint arXiv:1709.01507}}
  (\bibinfo{year}{2017}).
\newblock


\bibitem[\protect\citeauthoryear{Huang, Liu, Van Der~Maaten, and
  Weinberger}{Huang et~al\mbox{.}}{2017}]%
        {huang2017densely}
\bibfield{author}{\bibinfo{person}{Gao Huang}, \bibinfo{person}{Zhuang Liu},
  \bibinfo{person}{Laurens Van Der~Maaten}, {and} \bibinfo{person}{Kilian~Q
  Weinberger}.} \bibinfo{year}{2017}\natexlab{}.
\newblock \showarticletitle{Densely connected convolutional networks}. In
  \bibinfo{booktitle}{\emph{Proceedings of the IEEE conference on computer
  vision and pattern recognition}}. \bibinfo{pages}{4700--4708}.
\newblock


\bibitem[\protect\citeauthoryear{Kingma and Ba}{Kingma and Ba}{2014}]%
        {kingma2014adam}
\bibfield{author}{\bibinfo{person}{Diederik~P Kingma} {and}
  \bibinfo{person}{Jimmy Ba}.} \bibinfo{year}{2014}\natexlab{}.
\newblock \showarticletitle{Adam: A method for stochastic optimization}.
\newblock \bibinfo{journal}{\emph{arXiv preprint arXiv:1412.6980}}
  (\bibinfo{year}{2014}).
\newblock


\bibitem[\protect\citeauthoryear{Kingma and Welling}{Kingma and
  Welling}{2015}]%
        {kingma2015vae}
\bibfield{author}{\bibinfo{person}{Diederik~P Kingma} {and}
  \bibinfo{person}{Max Welling}.} \bibinfo{year}{2015}\natexlab{}.
\newblock \showarticletitle{Auto-encoding variational Bayes.}
\newblock \bibinfo{journal}{\emph{arXiv preprint arXiv:1312.6114}}
  (\bibinfo{year}{2015}).
\newblock


\bibitem[\protect\citeauthoryear{Krizhevsky, Sutskever, and Hinton}{Krizhevsky
  et~al\mbox{.}}{2012}]%
        {krizhevsky2012imagenet}
\bibfield{author}{\bibinfo{person}{Alex Krizhevsky}, \bibinfo{person}{Ilya
  Sutskever}, {and} \bibinfo{person}{Geoffrey~E Hinton}.}
  \bibinfo{year}{2012}\natexlab{}.
\newblock \showarticletitle{Imagenet classification with deep convolutional
  neural networks}. In \bibinfo{booktitle}{\emph{Advances in neural information
  processing systems}}. \bibinfo{pages}{1097--1105}.
\newblock


\bibitem[\protect\citeauthoryear{Li, Yumer, and Ramanan}{Li
  et~al\mbox{.}}{2020}]%
        {li2020budgeted}
\bibfield{author}{\bibinfo{person}{Mengtian Li}, \bibinfo{person}{Ersin Yumer},
  {and} \bibinfo{person}{Deva Ramanan}.} \bibinfo{year}{2020}\natexlab{}.
\newblock \bibinfo{title}{Budgeted Training: Rethinking Deep Neural Network
  Training Under Resource Constraints}.
\newblock
\newblock
\showeprint[arxiv]{1905.04753}~[cs.CV]


\bibitem[\protect\citeauthoryear{Li and Arora}{Li and Arora}{2020}]%
        {Li2020An}
\bibfield{author}{\bibinfo{person}{Zhiyuan Li} {and} \bibinfo{person}{Sanjeev
  Arora}.} \bibinfo{year}{2020}\natexlab{}.
\newblock \showarticletitle{An Exponential Learning Rate Schedule for Deep
  Learning}. In \bibinfo{booktitle}{\emph{International Conference on Learning
  Representations}}.
\newblock
\urldef\tempurl%
\url{https://openreview.net/forum?id=rJg8TeSFDH}
\showURL{%
\tempurl}


\bibitem[\protect\citeauthoryear{Liu, Jiang, He, Chen, Liu, Gao, and Han}{Liu
  et~al\mbox{.}}{2020}]%
        {liu2020variance}
\bibfield{author}{\bibinfo{person}{Liyuan Liu}, \bibinfo{person}{Haoming
  Jiang}, \bibinfo{person}{Pengcheng He}, \bibinfo{person}{Weizhu Chen},
  \bibinfo{person}{Xiaodong Liu}, \bibinfo{person}{Jianfeng Gao}, {and}
  \bibinfo{person}{Jiawei Han}.} \bibinfo{year}{2020}\natexlab{}.
\newblock \bibinfo{title}{On the Variance of the Adaptive Learning Rate and
  Beyond}.
\newblock
\newblock
\showeprint[arxiv]{1908.03265}~[cs.LG]


\bibitem[\protect\citeauthoryear{Loshchilov and Hutter}{Loshchilov and
  Hutter}{2017a}]%
        {loshchilov2017fixing}
\bibfield{author}{\bibinfo{person}{Ilya Loshchilov} {and}
  \bibinfo{person}{Frank Hutter}.} \bibinfo{year}{2017}\natexlab{a}.
\newblock \showarticletitle{Fixing weight decay regularization in adam}.
\newblock \bibinfo{journal}{\emph{arXiv preprint arXiv:1711.05101}}
  (\bibinfo{year}{2017}).
\newblock


\bibitem[\protect\citeauthoryear{Loshchilov and Hutter}{Loshchilov and
  Hutter}{2017b}]%
        {loshchilov2017sgdr}
\bibfield{author}{\bibinfo{person}{Ilya Loshchilov} {and}
  \bibinfo{person}{Frank Hutter}.} \bibinfo{year}{2017}\natexlab{b}.
\newblock \bibinfo{title}{SGDR: Stochastic Gradient Descent with Warm
  Restarts}.
\newblock
\newblock
\showeprint[arxiv]{1608.03983}~[cs.LG]


\bibitem[\protect\citeauthoryear{Lucas, Sun, Zemel, and Grosse}{Lucas
  et~al\mbox{.}}{2018}]%
        {lucas2018aggregated}
\bibfield{author}{\bibinfo{person}{James Lucas}, \bibinfo{person}{Shengyang
  Sun}, \bibinfo{person}{Richard Zemel}, {and} \bibinfo{person}{Roger Grosse}.}
  \bibinfo{year}{2018}\natexlab{}.
\newblock \showarticletitle{Aggregated momentum: Stability through passive
  damping}.
\newblock \bibinfo{journal}{\emph{arXiv preprint arXiv:1804.00325}}
  (\bibinfo{year}{2018}).
\newblock


\bibitem[\protect\citeauthoryear{O’donoghue and Candes}{O’donoghue and
  Candes}{2015}]%
        {o2015adaptive}
\bibfield{author}{\bibinfo{person}{Brendan O’donoghue} {and}
  \bibinfo{person}{Emmanuel Candes}.} \bibinfo{year}{2015}\natexlab{}.
\newblock \showarticletitle{Adaptive restart for accelerated gradient schemes}.
\newblock \bibinfo{journal}{\emph{Foundations of computational mathematics}}
  \bibinfo{volume}{15}, \bibinfo{number}{3} (\bibinfo{year}{2015}),
  \bibinfo{pages}{715--732}.
\newblock


\bibitem[\protect\citeauthoryear{Paszke, Gross, Chintala, Chanan, Yang, DeVito,
  Lin, Desmaison, Antiga, and Lerer}{Paszke et~al\mbox{.}}{2017}]%
        {paszke2017automatic}
\bibfield{author}{\bibinfo{person}{Adam Paszke}, \bibinfo{person}{Sam Gross},
  \bibinfo{person}{Soumith Chintala}, \bibinfo{person}{Gregory Chanan},
  \bibinfo{person}{Edward Yang}, \bibinfo{person}{Zachary DeVito},
  \bibinfo{person}{Zeming Lin}, \bibinfo{person}{Alban Desmaison},
  \bibinfo{person}{Luca Antiga}, {and} \bibinfo{person}{Adam Lerer}.}
  \bibinfo{year}{2017}\natexlab{}.
\newblock \showarticletitle{Automatic differentiation in PyTorch}.
\newblock  (\bibinfo{year}{2017}).
\newblock


\bibitem[\protect\citeauthoryear{Redmon, Divvala, Girshick, and Farhadi}{Redmon
  et~al\mbox{.}}{2016}]%
        {redmon2016look}
\bibfield{author}{\bibinfo{person}{Joseph Redmon}, \bibinfo{person}{Santosh
  Divvala}, \bibinfo{person}{Ross Girshick}, {and} \bibinfo{person}{Ali
  Farhadi}.} \bibinfo{year}{2016}\natexlab{}.
\newblock \bibinfo{title}{You Only Look Once: Unified, Real-Time Object
  Detection}.
\newblock
\newblock
\showeprint[arxiv]{1506.02640}~[cs.CV]


\bibitem[\protect\citeauthoryear{Redmon and Farhadi}{Redmon and
  Farhadi}{2018}]%
        {redmon2018yolov3}
\bibfield{author}{\bibinfo{person}{Joseph Redmon} {and} \bibinfo{person}{Ali
  Farhadi}.} \bibinfo{year}{2018}\natexlab{}.
\newblock \bibinfo{title}{YOLOv3: An Incremental Improvement}.
\newblock
\newblock
\showeprint[arxiv]{1804.02767}~[cs.CV]


\bibitem[\protect\citeauthoryear{Sanh, Debut, Chaumond, and Wolf}{Sanh
  et~al\mbox{.}}{2020}]%
        {sanh2020distilbert}
\bibfield{author}{\bibinfo{person}{Victor Sanh}, \bibinfo{person}{Lysandre
  Debut}, \bibinfo{person}{Julien Chaumond}, {and} \bibinfo{person}{Thomas
  Wolf}.} \bibinfo{year}{2020}\natexlab{}.
\newblock \bibinfo{title}{DistilBERT, a distilled version of BERT: smaller,
  faster, cheaper and lighter}.
\newblock
\newblock
\showeprint[arxiv]{1910.01108}~[cs.CL]


\bibitem[\protect\citeauthoryear{Shawahna, Sait, and El-Maleh}{Shawahna
  et~al\mbox{.}}{2019}]%
        {Shawahna_2019}
\bibfield{author}{\bibinfo{person}{Ahmad Shawahna}, \bibinfo{person}{Sadiq~M.
  Sait}, {and} \bibinfo{person}{Aiman El-Maleh}.}
  \bibinfo{year}{2019}\natexlab{}.
\newblock \showarticletitle{FPGA-Based Accelerators of Deep Learning Networks
  for Learning and Classification: A Review}.
\newblock \bibinfo{journal}{\emph{IEEE Access}}  \bibinfo{volume}{7}
  (\bibinfo{year}{2019}), \bibinfo{pages}{7823–7859}.
\newblock
\showISSN{2169-3536}
\urldef\tempurl%
\url{https://doi.org/10.1109/access.2018.2890150}
\showDOI{\tempurl}


\bibitem[\protect\citeauthoryear{Simonyan and Zisserman}{Simonyan and
  Zisserman}{2014}]%
        {simonyan2014very}
\bibfield{author}{\bibinfo{person}{Karen Simonyan} {and}
  \bibinfo{person}{Andrew Zisserman}.} \bibinfo{year}{2014}\natexlab{}.
\newblock \showarticletitle{Very deep convolutional networks for large-scale
  image recognition}.
\newblock \bibinfo{journal}{\emph{arXiv preprint arXiv:1409.1556}}
  (\bibinfo{year}{2014}).
\newblock


\bibitem[\protect\citeauthoryear{Smith}{Smith}{2018}]%
        {smith20181cycle}
\bibfield{author}{\bibinfo{person}{Leslie Smith}.}
  \bibinfo{year}{2018}\natexlab{}.
\newblock \showarticletitle{A disciplined approach to neural network
  hyper-parameters: Part 1 -- learning rate, batch size, momentum, and weight
  decay}.
\newblock \bibinfo{journal}{\emph{arXiv preprint arXiv:1803.09820}}
  (\bibinfo{year}{2018}).
\newblock


\bibitem[\protect\citeauthoryear{Smith}{Smith}{2017}]%
        {smith2017cyclical}
\bibfield{author}{\bibinfo{person}{Leslie~N. Smith}.}
  \bibinfo{year}{2017}\natexlab{}.
\newblock \bibinfo{title}{Cyclical Learning Rates for Training Neural
  Networks}.
\newblock
\newblock
\showeprint[arxiv]{1506.01186}~[cs.CV]


\bibitem[\protect\citeauthoryear{Smith, Kindermans, Ying, and Le}{Smith
  et~al\mbox{.}}{2017}]%
        {smith2017increasebatch}
\bibfield{author}{\bibinfo{person}{Samuel Smith}, \bibinfo{person}{Pieter-Jan
  Kindermans}, \bibinfo{person}{Chris Ying}, {and} \bibinfo{person}{Quoc Le}.}
  \bibinfo{year}{2017}\natexlab{}.
\newblock \showarticletitle{Don't Decay the Learning Rate, Increase the Batch
  Size}.
\newblock \bibinfo{journal}{\emph{arXiv preprint arXiv:1711.00489}}
  (\bibinfo{year}{2017}).
\newblock


\bibitem[\protect\citeauthoryear{Sonderby, Raiko, Maaloe, Sonderby, and
  Winther}{Sonderby et~al\mbox{.}}{2016}]%
        {sonderby2016ladder}
\bibfield{author}{\bibinfo{person}{Casper~Kaae Sonderby},
  \bibinfo{person}{Tapani Raiko}, \bibinfo{person}{Lars Maaloe},
  \bibinfo{person}{Soren~Kaae Sonderby}, {and} \bibinfo{person}{Ole Winther}.}
  \bibinfo{year}{2016}\natexlab{}.
\newblock \bibinfo{title}{Ladder Variational Autoencoders}.
\newblock
\newblock
\showeprint[arxiv]{1602.02282}~[stat.ML]


\bibitem[\protect\citeauthoryear{Sutskever, Martens, Dahl, and
  Hinton}{Sutskever et~al\mbox{.}}{2013}]%
        {sutskever2013importance}
\bibfield{author}{\bibinfo{person}{Ilya Sutskever}, \bibinfo{person}{James
  Martens}, \bibinfo{person}{George Dahl}, {and} \bibinfo{person}{Geoffrey
  Hinton}.} \bibinfo{year}{2013}\natexlab{}.
\newblock \showarticletitle{On the importance of initialization and momentum in
  deep learning}. In \bibinfo{booktitle}{\emph{International conference on
  machine learning}}. \bibinfo{pages}{1139--1147}.
\newblock


\bibitem[\protect\citeauthoryear{Sze, Chen, Emer, Suleiman, and Zhang}{Sze
  et~al\mbox{.}}{2017}]%
        {Sze_2017}
\bibfield{author}{\bibinfo{person}{Vivienne Sze}, \bibinfo{person}{Yu-Hsin
  Chen}, \bibinfo{person}{Joel Emer}, \bibinfo{person}{Amr Suleiman}, {and}
  \bibinfo{person}{Zhengdong Zhang}.} \bibinfo{year}{2017}\natexlab{}.
\newblock \showarticletitle{Hardware for machine learning: Challenges and
  opportunities}.
\newblock \bibinfo{journal}{\emph{2017 IEEE Custom Integrated Circuits
  Conference (CICC)}} (\bibinfo{date}{Apr} \bibinfo{year}{2017}).
\newblock
\showISBNx{9781509051915}
\urldef\tempurl%
\url{https://doi.org/10.1109/cicc.2017.7993626}
\showDOI{\tempurl}


\bibitem[\protect\citeauthoryear{Tripathi, Lipton, Belongie, and
  Nguyen}{Tripathi et~al\mbox{.}}{2016}]%
        {tripathi2016context}
\bibfield{author}{\bibinfo{person}{Subarna Tripathi},
  \bibinfo{person}{Zachary~C. Lipton}, \bibinfo{person}{Serge Belongie}, {and}
  \bibinfo{person}{Truong Nguyen}.} \bibinfo{year}{2016}\natexlab{}.
\newblock \bibinfo{title}{Context Matters: Refining Object Detection in Video
  with Recurrent Neural Networks}.
\newblock
\newblock
\showeprint[arxiv]{1607.04648}~[cs.CV]


\bibitem[\protect\citeauthoryear{Vahdat and Kautz}{Vahdat and Kautz}{2021}]%
        {vahdat2021nvae}
\bibfield{author}{\bibinfo{person}{Arash Vahdat} {and} \bibinfo{person}{Jan
  Kautz}.} \bibinfo{year}{2021}\natexlab{}.
\newblock \bibinfo{title}{NVAE: A Deep Hierarchical Variational Autoencoder}.
\newblock
\newblock
\showeprint[arxiv]{2007.03898}~[stat.ML]


\bibitem[\protect\citeauthoryear{Wolf, Debut, Sanh, Chaumond, Delangue, Moi,
  Cistac, Rault, Louf, Funtowicz, Davison, Shleifer, von Platen, Ma, Jernite,
  Plu, Xu, Scao, Gugger, Drame, Lhoest, and Rush}{Wolf et~al\mbox{.}}{2020}]%
        {wolf2020huggingfaces}
\bibfield{author}{\bibinfo{person}{Thomas Wolf}, \bibinfo{person}{Lysandre
  Debut}, \bibinfo{person}{Victor Sanh}, \bibinfo{person}{Julien Chaumond},
  \bibinfo{person}{Clement Delangue}, \bibinfo{person}{Anthony Moi},
  \bibinfo{person}{Pierric Cistac}, \bibinfo{person}{Tim Rault},
  \bibinfo{person}{Rémi Louf}, \bibinfo{person}{Morgan Funtowicz},
  \bibinfo{person}{Joe Davison}, \bibinfo{person}{Sam Shleifer},
  \bibinfo{person}{Patrick von Platen}, \bibinfo{person}{Clara Ma},
  \bibinfo{person}{Yacine Jernite}, \bibinfo{person}{Julien Plu},
  \bibinfo{person}{Canwen Xu}, \bibinfo{person}{Teven~Le Scao},
  \bibinfo{person}{Sylvain Gugger}, \bibinfo{person}{Mariama Drame},
  \bibinfo{person}{Quentin Lhoest}, {and} \bibinfo{person}{Alexander~M. Rush}.}
  \bibinfo{year}{2020}\natexlab{}.
\newblock \bibinfo{title}{HuggingFace's Transformers: State-of-the-art Natural
  Language Processing}.
\newblock
\newblock
\showeprint[arxiv]{1910.03771}~[cs.CL]


\bibitem[\protect\citeauthoryear{Yang and Shami}{Yang and Shami}{2020}]%
        {Yang_2020}
\bibfield{author}{\bibinfo{person}{Li Yang} {and} \bibinfo{person}{Abdallah
  Shami}.} \bibinfo{year}{2020}\natexlab{}.
\newblock \showarticletitle{On hyperparameter optimization of machine learning
  algorithms: Theory and practice}.
\newblock \bibinfo{journal}{\emph{Neurocomputing}}  \bibinfo{volume}{415}
  (\bibinfo{date}{Nov} \bibinfo{year}{2020}), \bibinfo{pages}{295–316}.
\newblock
\showISSN{0925-2312}
\urldef\tempurl%
\url{https://doi.org/10.1016/j.neucom.2020.07.061}
\showDOI{\tempurl}


\bibitem[\protect\citeauthoryear{Yeung, Kannan, Dauphin, and Fei-Fei}{Yeung
  et~al\mbox{.}}{2017}]%
        {yeung2017tackling}
\bibfield{author}{\bibinfo{person}{Serena Yeung}, \bibinfo{person}{Anitha
  Kannan}, \bibinfo{person}{Yann Dauphin}, {and} \bibinfo{person}{Li Fei-Fei}.}
  \bibinfo{year}{2017}\natexlab{}.
\newblock \bibinfo{title}{Tackling Over-pruning in Variational Autoencoders}.
\newblock
\newblock
\showeprint[arxiv]{1706.03643}~[cs.LG]


\bibitem[\protect\citeauthoryear{You, Gitman, and Ginsburg}{You
  et~al\mbox{.}}{2017}]%
        {you2017large}
\bibfield{author}{\bibinfo{person}{Yang You}, \bibinfo{person}{Igor Gitman},
  {and} \bibinfo{person}{Boris Ginsburg}.} \bibinfo{year}{2017}\natexlab{}.
\newblock \bibinfo{title}{Large Batch Training of Convolutional Networks}.
\newblock
\newblock
\showeprint[arxiv]{1708.03888}~[cs.CV]


\bibitem[\protect\citeauthoryear{Yuan, Ying, and Sayed}{Yuan
  et~al\mbox{.}}{2016}]%
        {yuan2016sgdequivalence}
\bibfield{author}{\bibinfo{person}{Kun Yuan}, \bibinfo{person}{Bicheng Ying},
  {and} \bibinfo{person}{Ali Sayed}.} \bibinfo{year}{2016}\natexlab{}.
\newblock \showarticletitle{On the influence of momentum acceleration on online
  learning}.
\newblock \bibinfo{journal}{\emph{Journal of Machine Learning Research}}
  \bibinfo{volume}{17}, \bibinfo{number}{192} (\bibinfo{year}{2016}),
  \bibinfo{pages}{1--66}.
\newblock


\bibitem[\protect\citeauthoryear{Zagoruyko and Komodakis}{Zagoruyko and
  Komodakis}{2016}]%
        {zagoruyko2016wide}
\bibfield{author}{\bibinfo{person}{Sergey Zagoruyko} {and}
  \bibinfo{person}{Nikos Komodakis}.} \bibinfo{year}{2016}\natexlab{}.
\newblock \showarticletitle{Wide residual networks}.
\newblock \bibinfo{journal}{\emph{arXiv preprint arXiv:1605.07146}}
  (\bibinfo{year}{2016}).
\newblock


\bibitem[\protect\citeauthoryear{Zeiler}{Zeiler}{2012}]%
        {zeiler2012adadelta}
\bibfield{author}{\bibinfo{person}{Matthew~D Zeiler}.}
  \bibinfo{year}{2012}\natexlab{}.
\newblock \showarticletitle{ADADELTA: an adaptive learning rate method}.
\newblock \bibinfo{journal}{\emph{arXiv preprint arXiv:1212.5701}}
  (\bibinfo{year}{2012}).
\newblock


\bibitem[\protect\citeauthoryear{Zhang and Mitliagkas}{Zhang and
  Mitliagkas}{2017}]%
        {zhang2017yellowfin}
\bibfield{author}{\bibinfo{person}{Jian Zhang} {and} \bibinfo{person}{Ioannis
  Mitliagkas}.} \bibinfo{year}{2017}\natexlab{}.
\newblock \showarticletitle{Yellowfin and the art of momentum tuning}.
\newblock \bibinfo{journal}{\emph{arXiv preprint arXiv:1706.03471}}
  (\bibinfo{year}{2017}).
\newblock


\end{thebibliography}



\end{document}